\PassOptionsToPackage{table}{xcolor}
\documentclass{article}
\usepackage{iclr2027_conference}
\usepackage{times}
\usepackage{natbib}
\usepackage{xcolor,colortbl,booktabs,graphicx}
\usepackage{amsmath,amssymb,amsthm}
\usepackage{enumitem}
\usepackage{array,tabularx}
\usepackage{longtable}
\usepackage{placeins}

\usepackage{pgfplots}
\pgfplotsset{compat=1.18}
\definecolor{RECAPPlotBlue}{RGB}{49,104,166}
\definecolor{RECAPPlotTeal}{RGB}{29,132,119}
\definecolor{RECAPPlotAmber}{RGB}{191,112,55}
\definecolor{RECAPPlotGray}{RGB}{108,116,128}
\pgfplotsset{
  recap/axis/.style={
    width=\linewidth,height=4.4cm,
    font=\fontsize{8}{9.5}\selectfont,
    title style={font=\fontsize{8.5}{10}\selectfont\bfseries},
    label style={font=\fontsize{8}{9.5}\selectfont},
    tick label style={font=\fontsize{8}{9.5}\selectfont},
    axis lines=left,axis line style={draw=black!45,line width=0.45pt},
    tick style={draw=black!40,line width=0.4pt},tick align=outside,
    major tick length=2pt,grid=major,
    grid style={draw=black!9,line width=0.3pt},
    scaled ticks=false,clip=true,
    legend style={draw=none,fill=white,font=\fontsize{8}{9.5}\selectfont,
      cells={anchor=west},inner sep=2pt},
    every axis plot/.append style={line width=1pt,mark size=2pt},
    /pgf/number format/1000 sep={,}
  },
  recap/candidate axis/.style={
    recap/axis,xmin=0.7,xmax=5.3,xtick={1,2,3,4,5},
    xlabel={Recorded candidate order},height=4.5cm
  }
}

\usepackage{microtype}
\usepackage{hyperref}
\usepackage{url}
\hypersetup{hidelinks}
\definecolor{BestCell}{RGB}{248,211,211}
\definecolor{SecondCell}{RGB}{250,226,194}
\definecolor{ThirdCell}{RGB}{255,246,197}
\definecolor{NeutralCell}{RGB}{240,241,243}
\definecolor{ModelHeader}{RGB}{76,88,103}
\newtheorem{proposition}{Proposition}
\newtheorem{theorem}{Theorem}

\newcommand{\Kact}{\mathsf{K}}
\newcommand{\Cact}{\mathsf{C}}
\newcommand{\Dact}{\mathsf{D}}
\newcommand{\Cost}{\mathrm{Cost}}
\newcommand{\ind}{\mathbf{1}}
\DeclareMathOperator*{\argmax}{arg\,max}
\newcommand{\SCfigure}[2][width=\linewidth]{%
  \IfFileExists{#2.pdf}{\includegraphics[#1]{#2.pdf}}{%
  \IfFileExists{#2.png}{\includegraphics[#1]{#2.png}}{%
    \PackageWarning{RECAP}{Missing external figure #2.pdf or #2.png}%
    \fbox{\parbox{0.92\linewidth}{\centering
    External figure not found.\\\texttt{\detokenize{#2.pdf}}}}}}}
\newcommand{\SCIfFigureExists}[3]{%
  \IfFileExists{#1.pdf}{#2}{%
    \IfFileExists{#1.png}{#2}{#3}}}
\newlength{\RECAPsavedtextfloatsep}
\title{When to Forget: Risk-Controlled Context Adaptation for Long-Horizon Agents}
\author{%
\textbf{Chenle Chen\textsuperscript{1}, Yangbo Wei\textsuperscript{2},
Zhijie Zheng\textsuperscript{2}, Zhenyu Su\textsuperscript{2},
Qifan Wang\textsuperscript{2},}\\
\textbf{Junhong Qian\textsuperscript{2}, Chao Yao\textsuperscript{3},
Chen Wu\textsuperscript{4}, Lei He\textsuperscript{4}}\\[4pt]
\textsuperscript{1}University of California, Los Angeles\quad
\textsuperscript{2}Shanghai Jiao Tong University\\
\textsuperscript{3}Arizona State University\quad
\textsuperscript{4}Eastern Institute of Technology, Ningbo\\[2pt]
\texttt{chenle@ucla.edu}%
}
\iclrfinalcopy
\begin{document}
\maketitle
\lhead{Preprint}

\begin{abstract}
Long-horizon agents learn from failure, but the value of a failure record can
change as execution progresses. We study when retaining, compressing, or
removing the same history block improves subsequent execution relative to
full-history KEEP. We introduce RECAP, a framework that combines matched
context interventions, execution-relational prediction, and independent
policy certification. Matched environment states and rollout seeds measure
success and continuation prompt-cost differences; prefix-only relations
between a failure block and the current execution propose promising edits.
Independent task bundles then select a policy subject to harmful-intervention
and non-positive-benefit risk constraints under a fixed measurement protocol.
A five-seed longitudinal case holds both the failure block and its summary
fixed: compression fails at an early state and becomes a cost-saving
intervention at a later state. Across five backbones, RECAP produces
positive paired-success saving estimates with backbone-specific
reliability--efficiency tradeoffs. On Hunyuan-A13B, it edits 43.24\% of retained
checkpoints while matching KEEP's observed 91.89\% continuation success and
reducing unconditional mean continuation prompt cost by 4,619.8 tokens
(2.13\%). The framework provides an execution-level reference for deciding
when to forget: retain history by its value to what comes next.
\end{abstract}

\section{Introduction}
\label{sec:introduction}
How long should a failure remain in an agent's memory? For tool-using
language-model agents, rejected candidates prevent repeated search, failed
checks reveal unmet constraints, and recovery traces preserve clues needed
to complete a task \citep{yao2022react,NEURIPS2023_1b44b878}. These records are
useful precisely because later decisions depend on them. However, as the
environment changes, dependencies are resolved, or subgoals are completed,
the same records may cease to affect the remaining execution while still
consuming prompt tokens at every model call.

The difficulty is that \emph{the value of a history record depends on what
the current execution still needs it for}. An old failure can encode an
active constraint; a semantically relevant record can become redundant
after a state change. Recency and similarity describe properties of the
record, but neither directly answers the intervention question:
\emph{what would happen to subsequent success and cost if this record were
compressed or removed now?} Prompt compression and memory management offer
mechanisms for reducing context
\citep{jiang2023llmlingua,jiang2024longllmlingua,packer2023memgpt}.
Execution-aware selection additionally requires a reference against which
the consequences of those edits can be measured.

\begin{figure}[t]
\centering
\includegraphics[width=\linewidth,height=0.50\textheight,keepaspectratio,draft=false]{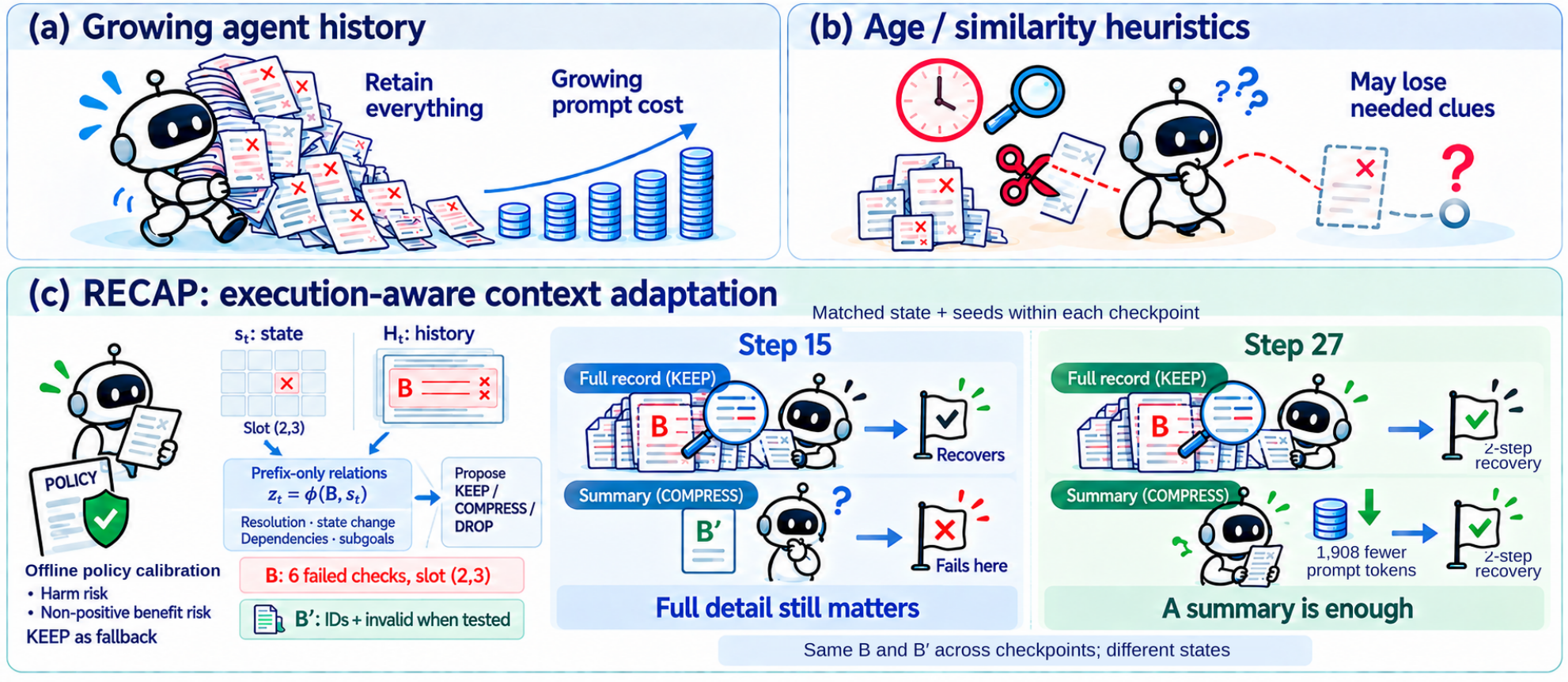}
\caption{\textbf{State-dependent value of the same failure block.}
(a--b) Schematic history growth and heuristic deletion.
(c) One fixed block $B$ and summary $B'$ in a Qwen3-8B course-selection task,
with five seeds and states matched across treatments within each checkpoint.
At step 15, KEEP succeeds and COMPRESS fails; at step 27, both recover in
two steps and COMPRESS saves 1,908 prompt tokens.
Figure~\ref{fig:same-block-longitudinal} compares all three treatments at seven checkpoints.}
\label{fig:state_dependent}
\end{figure}

Figure~\ref{fig:state_dependent} makes this distinction concrete. We evaluate
the same six failed slot checks and the same summary at two checkpoints of
one course-selection trajectory. At step 15, KEEP succeeds while COMPRESS
and DROP fail in all five seeds. At step 27, KEEP and COMPRESS recover in
two steps, but COMPRESS saves 1,908 continuation prompt tokens; DROP takes
12 recovery steps. Recovery denotes the first valid check of the target
slot, rather than task termination. The history block and its summary have
not changed; their measured value to the remaining execution has.

We formulate the decision as \emph{selective counterfactual improvement}
over KEEP. A finite-seed empirical oracle chooses a lower-cost edit only
when recorded success does not decrease, retaining the original otherwise.
RECAP (\emph{Risk-controlled Execution-aware Context Adaptation Policy})
approximates edit value using prefix-only information about relevance,
resolution, state changes, dependencies, and subgoals. It separates two
roles: learned scores propose candidate editing policies; tests on
independent calibration tasks select a policy whose two task-level risks
meet prespecified budgets. KEEP is the fallback when the evidence does not
support intervention. This separation instantiates Learn-Then-Test
\citep{angelopoulos2021learntest} for context adaptation rather than treating
predictive scores themselves as safety certificates.

Our contributions are:
\begin{itemize}
\item \textbf{An execution-based formulation of when to forget.}
Matched KEEP/COMPRESS/DROP continuations measure the success and cost effects
of changing one identified failure block at a fixed checkpoint.
\item \textbf{Execution-relational proposal with independent risk control.}
Prefix-only relations construct candidate policies, and independent
task-bundle tests select a rule under harmful-intervention and
non-positive-benefit constraints.
\item \textbf{State-dependent evidence and cross-backbone tradeoffs.}
A fixed-block longitudinal case isolates changing intervention value, while
a five-backbone, seven-selector comparison characterizes the resulting
reliability, savings, and intervention coverage.
\end{itemize}
The study evaluates one intervention per matched continuation. This gives a
controlled basis for learning and assessing an editing rule, with a
certificate for the declared task-bundle protocol.

\section{Related Work}
\label{sec:related}
\paragraph{Prompt compression and memory.}
Long-context performance can depend on where relevant evidence appears
\citep{liu2024lost}. Selective Context and the LLMLingua family reduce prompt
length through selective retention and question-aware or learned compression
\citep{li2023selectivecontext,jiang2023llmlingua,jiang2024longllmlingua,pan2024llmlingua2}.
Gist tokens and the In-context Autoencoder learn compact representations
\citep{mu2023gist,ge2024icae}. MemGPT manages memory tiers, A-MEM links notes,
and Mem0 consolidates conversational information
\citep{packer2023memgpt,xu2025amem,chhikara2025mem0}. These mechanisms address
how context is represented and stored. RECAP evaluates when a specified
failure-block edit is useful through its effect on subsequent execution.

\paragraph{Learned editing and experience reuse.}
ReAct, Reflexion, Self-Refine, and Voyager use interaction, feedback, or
accumulated experience to guide subsequent behavior
\citep{yao2022react,NEURIPS2023_1b44b878,madaan2023selfrefine,wang2023voyager}.
Context-Folding and AgentFold learn trajectory condensation, while AgeMem
learns memory operations
\citep{sun2025contextfolding,ye2025agentfold,yu2026agemem}.
ASSAY estimates per-skill contributions through randomized masking and
suppresses skills with negative predicted effects \citep{wang2026assay}.
RECAP focuses on failure records within an ongoing execution: the environment
is matched while the selected record is retained, compressed, or removed.
This directly connects an editing decision to a continuation-level reference.

\paragraph{Selection and statistical risk control.}
Selective prediction trades coverage for risk through abstention
\citep{geifman2017selective}. Risk-controlling prediction sets, Conformal Risk
Control, and Prompt Risk Control calibrate specified losses or prompt choices
\citep{bates2021rcps,angelopoulos2024crc,zollo2024promptrisk}.
We use Learn-Then-Test with one-sided Clopper--Pearson bounds
\citep{angelopoulos2021learntest,clopper1934binomial} to select from a
policy sequence fixed during development. KEEP provides abstention from
editing, and the two losses distinguish harmful success changes from
non-positive aggregate token benefit.

\paragraph{Baseline-relative improvement and evaluation.}
High-confidence off-policy evaluation and baseline-bootstrapped policy
improvement control deviations from a reference policy
\citep{thomas2015hcope,laroche2019spibb}. Our protocol directly evaluates all
three context treatments at each retained checkpoint, rather than estimating
full-policy returns from observational action logs. Experiments use the
configurable planning environments of AgentCE-Bench \citep{yang2026agentce}.
Extended memory and interactive-agent comparisons appear in
Appendix~\ref{app:related-work}.

\section{Single-Checkpoint Context Intervention}
\label{sec:problem}
\paragraph{Intervention unit.}
A checkpoint contains execution state $s$, ordered history $H$, and a
selected failure block $B\subset H$; we write $x=(s,B)$, with unchanged
history included in the checkpoint. The block records unsuccessful actions,
rejected candidates, or tool feedback from the available prefix. KEEP retains
$B$, COMPRESS replaces it with a deterministic summary, and DROP removes its
message pairs plus matching duplicate failure mentions in assistant text
(Appendix~\ref{app:block-transform}). Neither edit changes the true environment
state or actual past execution. All three treatments start from the same
checkpoint and differ only through these specified history transformations.
The backbone and treatment implementations remain frozen throughout a comparison.

\paragraph{Measured value.}
For matched seed $j$, let $Y_j(x,a)\in\{0,1\}$ be terminal success and
$\Cost_j(x,a)$ continuation prompt-token cost. The main comparison uses
$J=3$ seeds; the illustrative case uses five. Define
\begin{equation}
\begin{aligned}
\Delta_Y(x,a)&=\frac1J\sum_{j=1}^J[Y_j(x,a)-Y_j(x,\Kact)],\\
\Delta_C(x,a)&=\frac1J\sum_{j=1}^J[\Cost_j(x,\Kact)-\Cost_j(x,a)],\\
\mathcal A^+(x)&=\{\Kact\}\cup\{a\in\{\Cact,\Dact\}:\Delta_Y(x,a)\ge0,\ \Delta_C(x,a)>0\},\\
\pi^\star(x)&\in\argmax_{a\in\mathcal A^+(x)}\Delta_C(x,a).
\end{aligned}
\label{eq:oracle}
\end{equation}
A fixed tie rule completes the oracle. Positive $\Delta_C$ denotes a token
saving, and non-negative $\Delta_Y$ denotes no observed decrease in mean
success. KEEP has zero advantages relative to itself, so the admissible set
is always nonempty. Crucially, lower prompt length at the branch point is
not sufficient: $\Cost$ includes the remaining continuation, where repeated
search or additional recovery can offset the initial reduction.
The oracle is an offline finite-seed target; its outcomes are supervision
and evaluation data, not inputs available to the online selector.
The recovery-first preference diagnostic in Appendix~\ref{app:diagnostic-preference}
is reported separately from this token-based objective.

\paragraph{Matching protocol.}
A baseline trajectory identifies natural failures. Deterministic tool replay
restores each checkpoint without repeating prefix model generation. Tool
names, arguments, observations, state hashes, and block identity are checked
before the three treatments branch. The common seed grid aligns sampling
conditions while allowing subsequent trajectories to diverge. The same
frozen checkpoint--action matrix is reused for every selector: choosing KEEP
therefore reuses the identical reference outcome rather than a resampled
baseline. This protocol isolates the choice of history treatment from
changes in the agent, starting environment, or comparator rollout.

\paragraph{Execution-state dependence.}
Identical block text need not imply an identical best treatment. A
deterministic rule using only $B$ returns the same action at two checkpoints,
so it cannot be optimal at both if their optimal-action sets are disjoint
(Appendix~\ref{app:certificate}). Conditioning on execution allows the
selection rule to respond to what remains active, resolved, or relevant.
In the fixed-block study, both $B$ and its compressed replacement $B'$ are
held constant across checkpoints; only their execution context changes.

\section{RECAP: Proposal and Independent Certification}
\label{sec:method}
\begin{figure}[t]
\centering
\includegraphics[width=\linewidth,height=0.50\textheight,keepaspectratio,draft=false]{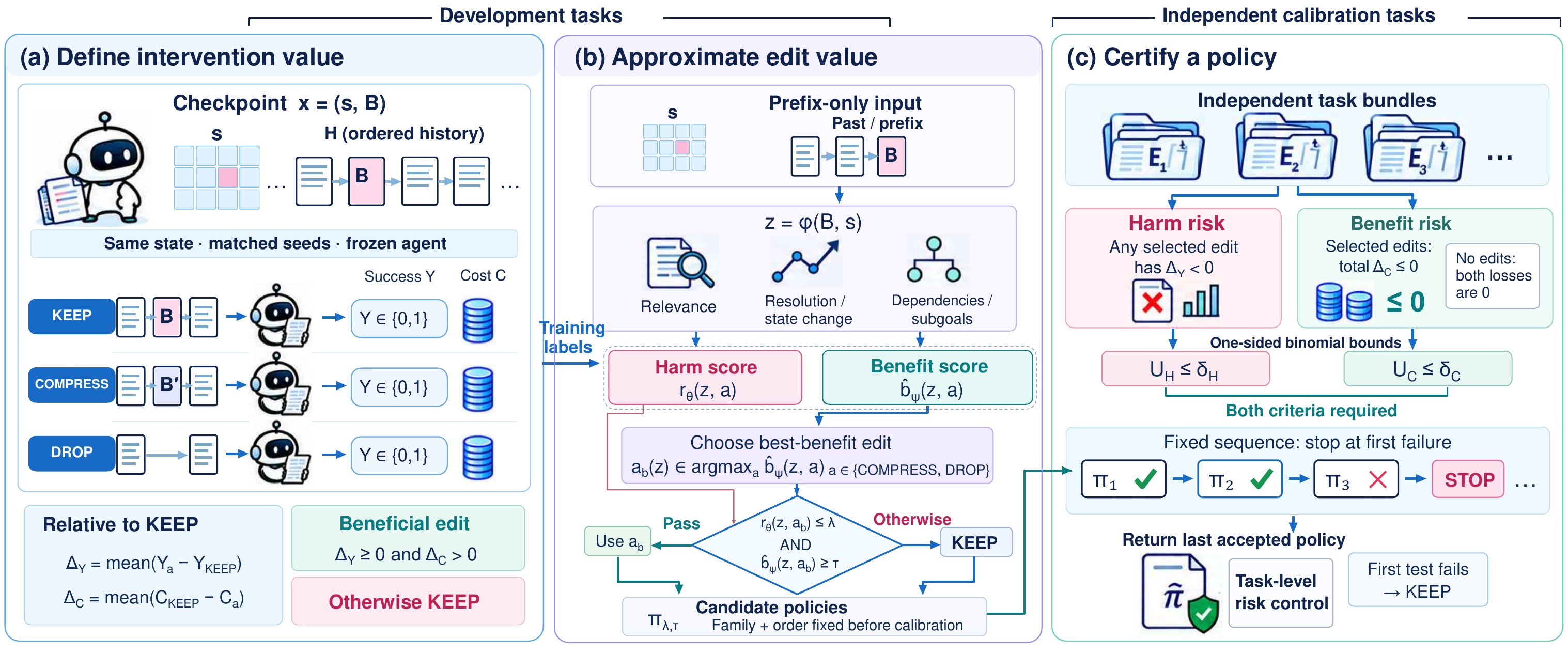}
\caption{\textbf{Measure, predict, and certify.}
(a) Matched continuations measure effects relative to KEEP.
(b) Prefix-only scores choose the highest-benefit edit; both gates must pass,
otherwise return KEEP.
(c) Independent task bundles test two policy-level risks. The candidate
family and order are fixed before calibration; stop at the first failure
and return the last accepted policy with calibration edits, or KEEP.
Certification is not a
per-edit or sequential-deployment guarantee.}
\label{fig:recap_overview}
\end{figure}

Figure~\ref{fig:recap_overview} separates measurement, prediction, and
certification. Development outcomes fit the scores and fix candidate
policies; independent calibration tasks select an admissible policy;
held-out outcomes evaluate that selected rule. The fitted predictors are
backbone-specific, while the intervention definitions and testing procedure
are common to all backbones. Appendix~\ref{app:operational} gives the
configuration table and operational protocol.

\subsection{Execution-relational proposal}
The prefix-only representation $z=\phi(B,s)$ contains 72 features in three
groups. \emph{Overlap and relevance} connect the selected record to the
current task and subgoal. \emph{Temporal and state-change relations}
describe execution progression and the status of the recorded failure,
including resolution or supersession. \emph{Dependencies} summarize active
paths and relationships to candidates and remaining subgoals. These groups
represent the relation of a record to the current execution, rather than
only the content of its text. Task/domain identifiers are excluded, and
standardization uses development tasks only. Two unscaled action indicators
are appended, giving 74 predictor inputs. Appendix~\ref{app:feature-schema}
defines all 72 state features. Future continuation outcomes are labels, not
feature inputs; the relation graph is not an identified causal graph.

For $a\in\{\Cact,\Dact\}$, the harm score $r_\theta(z,a)$ estimates the
recorded label $h(x,a)=\ind\{\Delta_Y(x,a)<0\}$. The benefit score
$\widehat b_\psi(z,a)$ regresses
$\widetilde{\Delta}_C(x,a)=\Delta_C(x,a)/\overline C_K(x)$, where
$\overline C_K(x)=J^{-1}\sum_j\Cost_j(x,\Kact)>0$. Thus $\tau$ is a score
threshold, not a raw-token budget. Predictor settings and the single-class
Hunyuan fallback are given in Appendix~\ref{app:predictor-config}. For each
development-fixed threshold pair $(\lambda,\tau)$, define
\begin{equation}
\begin{aligned}
a_b(z)&\in\argmax_{a\in\{\Cact,\Dact\}}\widehat b_\psi(z,a),\\
\pi_{\lambda,\tau}(z)&=
\begin{cases}
a_b(z),& r_\theta(z,a_b(z))\le\lambda\ \land\ \widehat b_\psi(z,a_b(z))\ge\tau,\\
\Kact,&\text{otherwise}.
\end{cases}
\end{aligned}
\label{eq:models-policy}
\end{equation}
The rule first selects the edit with the highest predicted benefit, then
requires both the harm and benefit gates to pass. Otherwise it returns KEEP
without trying the other edit. Equal predicted benefits prefer COMPRESS.
The gates serve distinct purposes: $\lambda$ screens predicted harm and
$\tau$ screens predicted benefit. Neither is a risk-penalty weight, and
neither replaces the independent policy test. The resulting candidate family
is $\Pi=\{\pi_{\lambda,\tau}:(\lambda,\tau)\in\Lambda\times\mathcal T\}$.

At a new checkpoint, the rule uses only the available prefix, frozen feature
preprocessing and predictors, and selected thresholds. Matched
KEEP/COMPRESS/DROP continuations construct development labels and calibration
losses; they are not evaluated as three live alternatives every time the
learned rule is invoked. The reference experiment and the resulting
score-based decision thus have distinct computational roles.

\subsection{Task-bundle risks}
A task bundle $E$ contains its retained checkpoints $X_E$ and all matched
outcomes. For $\pi(x)=\pi(\phi(B,s))$, define
\begin{equation}
\begin{aligned}
I_E(\pi)&=\{x\in X_E:\pi(x)\ne\Kact\},\qquad
D_E(\pi)=\sum_{x\in I_E(\pi)}\widetilde{\Delta}_C(x,\pi(x)),\\
\ell_H(E,\pi)&=\ind\{\exists x\in I_E(\pi):h(x,\pi(x))=1\},\\
\ell_C(E,\pi)&=\ind\{|I_E(\pi)|>0\ \land\ D_E(\pi)\le0\},\qquad
R_r(\pi)=\mathbb E_E[\ell_r(E,\pi)],\ r\in\{H,C\}.
\end{aligned}
\label{eq:task-risks}
\end{equation}
The harm loss is one if \emph{any} selected checkpoint has a measured success
decrease. It therefore does not average away a harmful decision using
improvements at other checkpoints. The cost loss instead aggregates the
selected edits' normalized continuation savings within a task and is one
when this aggregate is non-positive. Its sign need not match a raw-token sum
when KEEP costs differ across checkpoints. A policy can satisfy one criterion and fail the
other, making the two tests complementary. A task with no selected edit,
including an empty bundle, contributes zero losses but no intervention
coverage. Dependence among checkpoints within a task is allowed; independent
tasks, rather than individual continuations, are the calibration units.

The risks are expectations over the declared task-bundle population, not
conditional on making an edit. For a finite candidate family, the
coverage--risk tradeoff can be expressed as
\begin{equation}
\max_{\pi\in\Pi}\operatorname{Cov}(\pi)
\quad\text{subject to}\quad
R_H(\pi)\le\delta_H,\qquad R_C(\pi)\le\delta_C.
\label{eq:selection-objective}
\end{equation}
Coverage is the fraction of retained checkpoint decisions that differ from
KEEP. We implement a development-fixed ordered search through candidate
policies; Equation~\eqref{eq:selection-objective} states the population
objective rather than an optimization problem solved using test outcomes.

\subsection{Fixed-sequence policy selection}
Development fixes the predictors, preprocessing, threshold grid, ordered
family $(\pi_1,\ldots,\pi_M)$, and tie rules
(Appendix~\ref{app:candidate-construction}). For $m$ independent calibration
bundles, let $n_{\rm edit}(\pi)=\sum_i|I_{E_i}(\pi)|$ and define
\begin{equation}
\begin{aligned}
k_r(\pi)&=\sum_{i=1}^m\ell_r(E_i,\pi),\qquad
U_r(\pi)=U_{\rm CP}(k_r(\pi),m;\alpha/2),\\
\operatorname{pass}(\pi)&\Longleftrightarrow
U_H(\pi)\le\delta_H\ \land\ U_C(\pi)\le\delta_C,\\
\widehat q&=\max\bigl(\{0\}\cup\{q\in\{1,\ldots,M\}:n_{\rm edit}(\pi_q)>0,\\
&\hspace{8em}\operatorname{pass}(\pi_i)\ \forall i\le q\}\bigr),\\
\widehat\pi&=\pi_{\widehat q},\qquad \pi_0(x)\equiv\Kact.
\end{aligned}
\label{eq:certificate}
\end{equation}
Here $U_{\rm CP}$ is the one-sided Clopper--Pearson upper limit. Testing stops
at the first failed candidate and returns the last accepted candidate with
nonzero calibration edits, or KEEP if none exists. The threshold grid need not yield
nested policies or monotone risk/coverage. The experiments fix
$\alpha=0.05$ and $\delta_H=\delta_C=0.20$ before calibration, with one-sided
tail probability $\alpha/2=0.025$ per risk. Candidate order is not revised to
maximize observed calibration saving.

\begin{theorem}[Certification under the fixed measurement protocol]
Condition on a development-fixed ordered family. If calibration task bundles
are independent and identically distributed under the same fixed sampling
and measurement protocol as future task bundles, then
Equation~\eqref{eq:certificate} returns a policy satisfying
$R_H(\widehat\pi)\le\delta_H$ and $R_C(\widehat\pi)\le\delta_C$
with probability at least $1-\alpha$ over calibration.
\label{thm:certificate}
\end{theorem}
\begin{proof}[Proof sketch]
If the family contains an inadmissible candidate, consider the first one.
Selecting any inadmissible policy requires this candidate to pass, which
requires a false acceptance for at least one violated risk. The one-sided
exact test and a conservative union bound over the two coordinates bound
this probability by $\alpha$. KEEP has identically zero losses.
Appendix~\ref{app:certificate} gives the full argument.
\end{proof}
The guarantee controls the task-level events defined by the fixed,
finite-seed measurement protocol, separately for each backbone. It does not
require the predictive scores to be calibrated probabilities; statistical
validity comes from the independent task-level tests
\citep{angelopoulos2021learntest,clopper1934binomial}.

\section{Experiments}
\label{sec:experiments}
\subsection{Protocol, comparisons, and metrics}
We use AgentCE-Bench's course, meal, PC-build, shopping, travel, and workforce
planning domains \citep{yang2026agentce}. Five frozen backbones are evaluated:
Qwen3-4B, Qwen3-8B, Qwen3-14B (AWQ 4-bit), gpt-oss-20b (MXFP4), and
Hunyuan-A13B-Instruct (FP8). Each uses a separately fitted and calibrated
policy on its own screened task pool. Screening retains successful baseline
trajectories with at least two natural failures, followed by checkpoint
retention. The main Qwen3-8B split is 35/58/40 development/calibration/test
tasks. The 39-task diagnostic and standalone 69-task calibration are
separate studies; Appendix~\ref{app:reproducibility} maps all splits and
frozen policy identifiers.

The seven selectors are KEEP, Always-COMPRESS, Always-DROP, RECENCY,
SEMANTIC, LEARNED-NOCERT, and RECAP. They share treatment implementations
and a frozen outcome matrix, comparing selection rather than compressor
quality. On Hunyuan, RECENCY compresses above a development-selected age
threshold; SEMANTIC drops below a development-selected frozen MiniLM cosine
similarity threshold. LEARNED-NOCERT uses positive predicted benefit and a
harm score below $0.5$, without certification. RECAP chooses between both
edits using Equation~\eqref{eq:models-policy}. Full heuristic settings are
in Appendix~\ref{app:hunyuan-results}.

Continuation success averages terminal success over retained checkpoints and
seeds; switching is the fraction of checkpoint decisions that are non-KEEP.
Task harm averages $\ell_H$ over the stated task population. For Hunyuan,
the primary risk uses registered tasks in calibration (30) and test (19);
15 test tasks supply 37 checkpoints, and the remaining four have empty
bundles. Conditional retained-task rates are given separately in
Appendix~\ref{app:hunyuan-results}.
Unconditional saving $S_{\rm all}$ averages KEEP-minus-policy prompt cost over
all checkpoint--seed pairs, including failures and zero-valued KEEP decisions.
Relative saving $S_{\rm rel}=100S_{\rm all}/\overline C_K$ uses the corresponding
unconditional KEEP mean. Paired-success saving $S_{\rm pair}$ averages over
jointly successful pairs, whose subset can differ by selector. We report
both cost views where available, with full definitions in
Appendix~\ref{app:diagnostic-preference}.

\subsection{Held-out reliability--efficiency tradeoffs}
\label{sec:main-results}
Tables~\ref{tab:main-multimodel}--\ref{tab:main-crossfamily} report all seven
selectors. Across the five backbones, RECAP has positive paired-success
saving estimates, with observed continuation success within 1.6 percentage
points of KEEP. Always-COMPRESS and Always-DROP reduce observed success on
every backbone, showing why the cost and reliability effects of editing
must be considered jointly. Selective policies expose different operating
points between preserving the full history and editing every checkpoint.

\begin{table}[t]
\centering
\caption{\textbf{Held-out matched continuations: Qwen backbones.}
Success, task harm, and switching are percentages. $S_{\rm pair}$ is
prompt-token saving over jointly successful checkpoint--seed pairs,
including KEEP decisions; its subset can differ by selector.
Colors denote within-backbone ranks, not significance. Dashes are unreported.
Splits and thresholds are in Appendix~\ref{app:reproducibility}.}
\label{tab:main-multimodel}
\small
\setlength{\tabcolsep}{4pt}
\renewcommand{\arraystretch}{1.04}
\begin{tabular*}{\linewidth}{@{\extracolsep{\fill}}lrrrr@{}}
\toprule
Method & Success $\uparrow$ & Task harm $\downarrow$ & $S_{\rm pair}\uparrow$ & Switch \\
\midrule
\rowcolor{ModelHeader}\multicolumn{5}{l}{\textcolor{white}{\textbf{Qwen3-4B}}} \\
KEEP & \cellcolor{BestCell}96.3 & \cellcolor{NeutralCell}0.0 & \cellcolor{NeutralCell}0.0 & 0.00 \\
Always-COMPRESS & 81.5 & 47.6 & -9,488.0 & 100.00 \\
Always-DROP & 73.0 & 52.4 & -11,605.0 & 100.00 \\
RECENCY & \cellcolor{SecondCell}95.2 & \cellcolor{BestCell}4.8 & \cellcolor{ThirdCell}-548.0 & \textemdash \\
SEMANTIC & 88.4 & \cellcolor{ThirdCell}14.3 & -1,352.0 & \textemdash \\
LEARNED-NOCERT & 94.2 & \cellcolor{SecondCell}9.5 & \cellcolor{SecondCell}-250.0 & \textemdash \\
\textbf{RECAP} & \cellcolor{ThirdCell}94.7 & \cellcolor{BestCell}4.8 & \cellcolor{BestCell}+10.7 & 4.76 \\
\midrule
\rowcolor{ModelHeader}\multicolumn{5}{l}{\textcolor{white}{\textbf{Qwen3-8B}}} \\
KEEP & \cellcolor{BestCell}95.5 & \cellcolor{NeutralCell}0.0 & \cellcolor{NeutralCell}0.0 & 0.00 \\
Always-COMPRESS & 80.8 & 37.5 & -4,898.0 & 100.00 \\
Always-DROP & 90.1 & 27.5 & -2,319.0 & 100.00 \\
RECENCY & \cellcolor{BestCell}95.5 & \cellcolor{BestCell}0.0 & -1,719.0 & 25.42 \\
SEMANTIC & \cellcolor{ThirdCell}93.8 & \cellcolor{ThirdCell}10.0 & \cellcolor{ThirdCell}-698.0 & 44.92 \\
LEARNED-NOCERT & \cellcolor{BestCell}95.5 & \cellcolor{SecondCell}2.5 & \cellcolor{BestCell}+58.0 & 37.29 \\
\textbf{RECAP} & \cellcolor{SecondCell}94.6 & \cellcolor{SecondCell}2.5 & \cellcolor{SecondCell}+32.9 & 31.36 \\
\midrule
\rowcolor{ModelHeader}\multicolumn{5}{l}{\textcolor{white}{\textbf{Qwen3-14B (AWQ 4-bit)}}} \\
KEEP & \cellcolor{BestCell}97.5 & \cellcolor{NeutralCell}0.0 & \cellcolor{NeutralCell}0.0 & 0.00 \\
Always-COMPRESS & 89.1 & 31.7 & -4,716.0 & 100.00 \\
Always-DROP & 91.3 & 24.4 & -1,703.0 & 100.00 \\
RECENCY & \cellcolor{BestCell}97.5 & \cellcolor{BestCell}0.0 & \cellcolor{ThirdCell}+873.0 & 18.03 \\
SEMANTIC & \cellcolor{ThirdCell}96.2 & \cellcolor{ThirdCell}4.9 & \cellcolor{SecondCell}+1,306.0 & 27.87 \\
LEARNED-NOCERT & \cellcolor{SecondCell}97.3 & \cellcolor{SecondCell}2.4 & \cellcolor{BestCell}+1,847.0 & 45.08 \\
\textbf{RECAP} & \cellcolor{BestCell}97.5 & \cellcolor{BestCell}0.0 & +801.0 & 20.49 \\
\bottomrule
\end{tabular*}
\end{table}

\begin{table}[t]
\centering
\caption{\textbf{Held-out matched continuations: cross-family results.}
Success, task harm, and switching are percentages. Hunyuan harm uses all
19 registered test tasks, matching the calibration population convention;
15 supply checkpoints. $S_{\rm all}$ includes all continuations, $S_{\rm pair}$
only joint-success pairs, and $S_{\rm rel}=100S_{\rm all}/\overline C_K$.
Hunyuan $S_{\rm all}$ and $S_{\rm rel}$ are derived from the recorded mean
costs, not new rollouts; dashes are unavailable. Both savings include KEEP decisions.}
\label{tab:main-crossfamily}
\small
\setlength{\tabcolsep}{2.5pt}
\renewcommand{\arraystretch}{1.04}
\begin{tabular*}{\linewidth}{@{\extracolsep{\fill}}lrrrrrr@{}}
\toprule
Method & Success & Harm & Switch & $S_{\rm all}$ & $S_{\rm pair}$ & $S_{\rm rel}$ \\
\midrule
\rowcolor{ModelHeader}\multicolumn{7}{l}{\textcolor{white}{\textbf{gpt-oss-20b (MXFP4)}}} \\
KEEP & 73.9 & 0.0 & 0.00 & 0.0 & 0.0 & \textemdash \\
Always-COMPRESS & 67.7 & 44.4 & 100.00 & \textemdash & -22,191.0 & \textemdash \\
Always-DROP & 72.6 & 31.1 & 100.00 & \textemdash & +1,835.0 & \textemdash \\
RECENCY & 73.9 & 0.0 & \textemdash & \textemdash & -5,788.0 & \textemdash \\
SEMANTIC & 74.9 & 2.2 & \textemdash & \textemdash & +5,372.0 & \textemdash \\
LEARNED-NOCERT & 73.1 & 2.2 & \textemdash & \textemdash & +2,247.0 & \textemdash \\
\textbf{RECAP} & 73.9 & 0.0 & 2.33 & \textemdash & +182.5 & \textemdash \\
\midrule
\rowcolor{ModelHeader}\multicolumn{7}{l}{\textcolor{white}{\textbf{Hunyuan-A13B-Instruct (FP8)}}} \\
KEEP & 91.9 & 0.0 & 0.00 & 0.0 & 0.0 & 0.00\% \\
Always-COMPRESS & 88.3 & 15.8 & 100.00 & +11,614.2 & +8,700.6 & 5.35\% \\
Always-DROP & 90.1 & 5.3 & 100.00 & +19,296.5 & +20,235.7 & 8.88\% \\
RECENCY & 91.9 & 0.0 & 10.81 & -263.0 & -300.6 & -0.12\% \\
SEMANTIC & 90.1 & 5.3 & 54.05 & +14,387.5 & +14,969.9 & 6.62\% \\
LEARNED-NOCERT & 91.9 & 0.0 & 43.24 & +4,619.8 & +5,000.4 & 2.13\% \\
\textbf{RECAP} & 91.9 & 0.0 & 43.24 & +4,619.8 & +5,000.4 & 2.13\% \\
\bottomrule
\end{tabular*}
\end{table}

\paragraph{Hunyuan: selective editing with unchanged observed success.}
RECAP edits 16 of 37 checkpoints (43.24\%): ten COMPRESS and six DROP choices.
It matches KEEP's 91.89\% observed continuation success, with zero measured
harmful tasks under both registered and retained test denominators. Its 64
development edit labels contain no harm, so the frozen harm score is constant
$1/130$; state-dependent selection here comes from the benefit predictor,
not learned harm discrimination (Appendix~\ref{app:predictor-config}). Mean
prompt cost decreases from 217,219.0 to 212,599.2 tokens, giving an
unconditional saving of 4,619.8 per continuation (2.13\%). On the 102
jointly successful pairs, the mean saving is 5,000.4 tokens. Both metrics
include unchanged KEEP decisions and describe different averaging sets.
Always-DROP achieves a larger unconditional reduction of 8.88\%, but lowers
observed success and incurs one harmful task. SEMANTIC occupies another
tradeoff point, with 6.62\% unconditional saving and one harmful task.

\paragraph{Backbone-specific operating points.}
On Qwen3-4B and Qwen3-8B, RECAP's paired savings are +10.7 and +32.9 tokens,
with success 94.7\% and 94.6\%, respectively, compared with KEEP's 96.3\% and
95.5\%. On Qwen3-14B, RECAP preserves the observed 97.5\% success and saves
801 paired tokens while editing 20.49\% of checkpoints. gpt-oss selects
three DROP edits among 129 checkpoints (2.33\%), preserving 73.9\% observed
success and saving 182.5 paired tokens. These differences motivate fitting
and calibrating the selector for the backbone on which it will be used.
The Qwen-to-Hunyuan transfer diagnostic provides a corresponding example.
A Qwen-derived policy edits one of 37 Hunyuan checkpoints (2.70\%) and saves
5.9 paired-success tokens. Refitting and recalibrating for Hunyuan raise
coverage to 43.24\% and paired saving to 5,000.4, with unchanged observed
success on that matrix. This comparison evaluates the two adaptation steps
together. It connects the cross-backbone variation to the practical choice
of fitting and calibrating a rule for its target backbone rather than
reusing a single operating point across agent families.

\paragraph{Prediction, calibration, and test.}
The selected Hunyuan policy has $(\lambda,\tau)=(0.0076923077,0)$. On 30
registered calibration tasks, $k_H=0$ and $k_C=1$ give $U_H=0.1157$ and
$U_C=0.1722$, both below the 0.20 budgets. It selects ten of 41 calibration
checkpoints (K/C/D $=31/7/3$); the frozen rule is then evaluated on test,
where K/C/D $=21/10/6$. The candidate ledger in
Appendix~\ref{app:hunyuan-results} shows why this is policy selection under
risk constraints rather than maximizing calibration-sample savings.

The calibration ledger contains five distinct threshold pairs in the fixed
order $\tau=(0.10,0.05,0.01,0.02,0)$, with the same $\lambda$; nine repeated
quantile labels per pair account for its 45 rows. All five pairs pass both
tests, so the final pair is selected. Across the distinct candidates,
coverage ranges from two to ten edited checkpoints out of 41. The selected
candidate has the highest observed coverage but not the largest
calibration-sample saving. This illustrates the role of the predefined
search order: calibration checks admissibility of an editing rule, while
the held-out matrix assesses its realized reliability and efficiency.

Certification and empirical performance play different roles. On Hunyuan,
RECAP and LEARNED-NOCERT have equal reported success, cost, and action-count
aggregates. On Qwen3-14B, RECENCY matches the displayed success and zero task
harm while saving 873 rather than 801 paired tokens; on Qwen3-8B,
LEARNED-NOCERT has the same task harm and higher observed success and saving.
These comparisons locate the methods' operating points. The certificate
adds an independent test of the specified risks to the selected policy,
rather than an ordering of all selectors by test performance.

\subsection{State dependence and representation diagnostics}
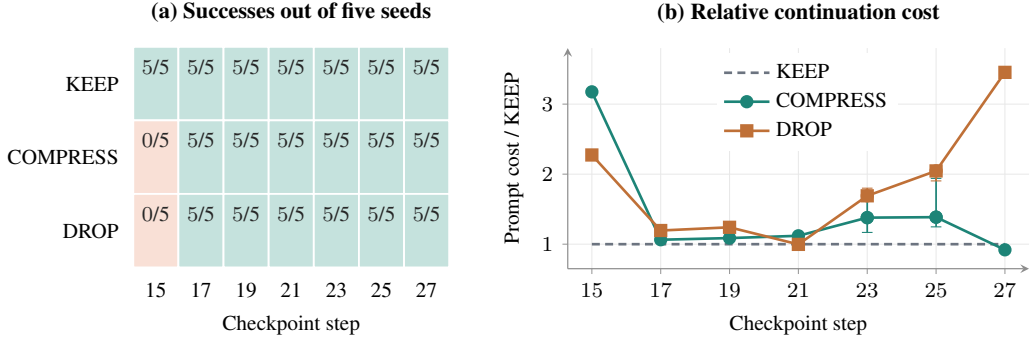
\begin{figure}[tbp]
\centering
\begin{minipage}[t]{0.43\linewidth}
\centering
\begin{tikzpicture}
\begin{axis}[
  recap/axis,height=4.5cm,width=\dimexpr\linewidth-7pt\relax,
  title={(a) Successes out of five seeds},
  xmin=0.5,xmax=7.5,ymin=-0.5,ymax=2.5,y dir=reverse,
  xtick={1,2,3,4,5,6,7},xticklabels={15,17,19,21,23,25,27},
  ytick={0,1,2},yticklabels={KEEP,COMPRESS,DROP},
  xlabel={Checkpoint step},grid=none,
  axis line style={draw=none},tick style={draw=none},
  point meta min=0,point meta max=5,
  colormap={recapSuccess}{rgb255(0cm)=(249,224,214); rgb255(1cm)=(197,226,221)}
]
\addplot[
  matrix plot*,mesh/cols=7,mesh/ordering=x varies,
  point meta=explicit,draw=white,line width=0.7pt,
  nodes near coords={\pgfmathprintnumber[fixed,precision=0]{\pgfplotspointmeta}/5},
  every node near coord/.append style={font=\fontsize{8}{9}\selectfont,text=black!85}
] table[x=x,y=y,meta=success,row sep=\\] {
x y success \\
1 0 5 \\
2 0 5 \\
3 0 5 \\
4 0 5 \\
5 0 5 \\
6 0 5 \\
7 0 5 \\
1 1 0 \\
2 1 5 \\
3 1 5 \\
4 1 5 \\
5 1 5 \\
6 1 5 \\
7 1 5 \\
1 2 0 \\
2 2 5 \\
3 2 5 \\
4 2 5 \\
5 2 5 \\
6 2 5 \\
7 2 5 \\
};
\end{axis}
\end{tikzpicture}
\end{minipage}\hfill
\begin{minipage}[t]{0.55\linewidth}
\centering
\begin{tikzpicture}
\begin{axis}[
  recap/axis,height=4.5cm,
  title={(b) Relative continuation cost},
  xmin=14.3,xmax=27.7,ymin=0.65,ymax=3.8,
  xtick={15,17,19,21,23,25,27},ytick={1,2,3},
  xlabel={Checkpoint step},ylabel={Prompt cost / KEEP},
  legend style={at={(0.52,0.98)},anchor=north},legend columns=1
]
\addplot[RECAPPlotGray,densely dashed,mark=none] coordinates {(15,1) (27,1)};
\addlegendentry{KEEP}
\addplot[RECAPPlotTeal,mark=*] coordinates {(15,3.17534406) (17,1.0636961) (19,1.08731659) (21,1.11985138) (23,1.37910941) (25,1.38597092) (27,0.919676686)};
\addlegendentry{COMPRESS}
\addplot[RECAPPlotAmber,mark=square*] coordinates {(15,2.27383951) (17,1.19522476) (19,1.24121583) (21,0.997275301) (23,1.69097669) (25,2.04641004) (27,3.45356572)};
\addlegendentry{DROP}
\addplot[RECAPPlotTeal,only marks,mark=none,forget plot,
  error bars/y dir=both,error bars/y explicit,
  error bars/error bar style={line width=0.65pt}]
  table[x=step,y=ratio,y error plus=plus,y error minus=minus,row sep=\\] {
step ratio plus minus \\
23 1.379109410 0.316620282 0.211080188 \\
25 1.385970915 0.553635586 0.138408896 \\
};
\addplot[RECAPPlotAmber,only marks,mark=none,forget plot,
  error bars/y dir=both,error bars/y explicit,
  error bars/error bar style={line width=0.65pt}]
  table[x=step,y=ratio,y error plus=plus,y error minus=minus,row sep=\\] {
step ratio plus minus \\
23 1.690976692 0.108836319 0.027209080 \\
25 2.046410037 0.035682920 0.142731679 \\
};
\end{axis}
\end{tikzpicture}
\end{minipage}
\caption{\textbf{One fixed failure block across seven execution states.}
A Qwen3-8B course-selection case with five seeds per checkpoint--action pair
(105 continuations). Left: observed success counts. Right: mean continuation
prompt cost relative to same-checkpoint KEEP, including failed runs; whiskers
show seed minima and maxima, not confidence intervals. Both $B$ and $B'$
are fixed. Compression preserves success from step 17 onward, but becomes
less costly than KEEP only at step 27 among these measured checkpoints.}
\label{fig:same-block-longitudinal}
\end{figure}

Figure~\ref{fig:same-block-longitudinal} expands the motivating example to
all seven checkpoints of the same Qwen3-8B trajectory. The fixed block
contains six failed checks for slot $(2,3)$, and its fixed summary retains
the tested candidate identifiers and their invalidity when tested. Holding
both representations constant separates changing execution value from
changes in summary content. The experiment contains five seeds for each
checkpoint--action pair, giving 105 continuations.

At step 15, KEEP succeeds in every seed, while both edits fail despite
shortening the first prompt. From step 17 onward all three treatments
succeed in the recorded seeds, but their continuation costs differ.
COMPRESS remains more costly than KEEP at the measured checkpoints through
step 25; at step 27 it reduces mean cost from 23,754 to 21,846 tokens
(8.03\%). DROP at that checkpoint costs 3.45 times KEEP. Success preservation
alone therefore does not identify a useful compression point.
The transition is also not monotone: at step 21, DROP saves 154 prompt
tokens but takes two more recovery steps. The token-based oracle selects
DROP there, whereas the recovery-first diagnostic selects KEEP. This case
illustrates why the objective, as well as the execution state, must be
specified when assigning value to a failure record.

A complementary development diagnostic contains 39 tasks, 286 states, and
2,574 matched rollouts. Its recovery-first preferences vary within 38 tasks;
90 states show success differences across treatments and the remaining 196
show efficiency differences with equal success. This broader diagnostic
does not hold the failure block fixed, unlike the longitudinal case above.
For harm prediction, execution-relational features achieve ROC--AUC 0.7015
versus 0.6147 for structured features using logistic regression in both
cases. A relational random forest achieves 0.7783 for token-beneficial
intervention prediction. These diagnostics support using execution relations
to score intervention effects. The semantic-plus-structured comparison uses
a different classifier and is reported separately with all estimator choices
in Appendix~\ref{app:development-diagnostics}.
The leave-one-task-out diagnostic evaluates effect prediction across tasks,
whereas the calibration procedure evaluates the losses of a complete
editing policy. The two analyses therefore address the proposal and
certification stages at their respective units of evaluation.

\section{Discussion}
\label{sec:limitations}
RECAP separates measured edit effects, prefix-based proposals, and task-level
policy certification, with KEEP as the common reference. Evaluation uses
one edit per matched continuation on screened tasks; checkpoint savings are
not episode-level savings. Certification assumes iid task bundles under the
same sampling and finite-seed measurement protocol. Registered-task and
retained-task risks use different denominators. The benefit test controls
the probability of non-positive aggregate normalized benefit, not loss
magnitude or expected raw-token saving. Reported efficiency is continuation
prompt cost; summaries use deterministic templates, and feature extraction,
text transformation, and controller runtime overhead are not separately itemized.

\section{Conclusion}
\label{sec:conclusion}
Failure history should be assessed by its contribution to subsequent
execution, rather than only by its age or textual relevance. RECAP makes
this principle operational through matched context interventions,
execution-relational proposals, and independent certification of two
task-bundle risks. A fixed-block longitudinal case demonstrates changing
intervention value, and a five-backbone evaluation identifies
backbone-specific reliability--efficiency tradeoffs. Hunyuan-A13B illustrates
selective editing with unchanged observed continuation success and a 2.13\%
unconditional prompt-cost reduction. Together, the measurement protocol
and policy-selection rule provide a concrete basis for deciding when
retention remains useful and when a shorter representation is sufficient.
\label{sec:main-end}
\clearpage
\setlength{\textfloatsep}{\RECAPsavedtextfloatsep}

\section*{Statement on the use of AI tools}
Generative AI tools assisted with language polishing, manuscript editing,
reference organization, LaTeX debugging, and the preparation of figures
and data-analysis code. The authors retain full responsibility for the
methods, proofs, experimental results, references, and final manuscript.

\section*{Reproducibility statement}
Sections~\ref{sec:problem}--\ref{sec:method} define the intervention unit,
measured targets, policy gates, and certification assumptions.
Appendix~\ref{app:operational} specifies the operational protocol and
Appendix~\ref{app:certificate} provides the complete proof.
Appendices~\ref{app:hunyuan-results}, \ref{app:replay}, and
\ref{app:reproducibility} document cost and calibration ledgers, the worked
fixed-block case, task splits, and frozen policy identifiers.
Appendix~\ref{app:implementation-status} specifies the feature dictionary,
block transformations, training targets, model construction, and candidate order.

\bibliographystyle{iclr2027_conference}
\bibliography{references}
\clearpage
\appendix
\numberwithin{table}{section}
\numberwithin{figure}{section}

\FloatBarrier
\section{Method Configuration and Operational Protocol}
\label{app:operational}
Table~\ref{tab:method-config} consolidates the recorded framework settings.
The following procedures specify the order of feature computation, action
selection, and independent calibration.

\begin{table}[htbp]
\centering
\caption{\textbf{Method configuration and evaluation protocol.}
Backbone-specific splits and thresholds appear in Table~\ref{tab:splits}.
The fixed-block example and implementation scope are described in
Appendix~\ref{app:replay}.}
\label{tab:method-config}
\small
\setlength{\tabcolsep}{5pt}
\renewcommand{\arraystretch}{1.08}
\begin{tabularx}{\linewidth}{@{}>{\raggedright\arraybackslash}p{0.19\linewidth}>{\raggedright\arraybackslash}X@{}}
\toprule
Component & Configuration \\
\midrule
Checkpoint and block & Successful screening trajectory with at least two
natural failures; retained checkpoint $x=(s,B)$ and selected failure block.
Selection is fixed before comparing treatments. \\
Context treatments & KEEP retains $B$; COMPRESS uses a deterministic
JSON/text template. DROP removes selected message pairs and matching duplicate
failure mentions in assistant text. The environment is unchanged. \\
Matched measurement & Three rollout seeds per checkpoint and action
(five in the illustrative study); terminal success and continuation prompt
cost; state/block hashes and replayed tool outputs checked before branching. \\
Relational input & 72 prefix-only features (43 overlap, 14 temporal, 15 graph),
plus two unscaled action indicators. Development-only standardization;
full dictionary in Appendix~\ref{app:feature-schema}. \\
Predictors & Logistic harm score and 100-tree normalized-benefit regressor;
Hunyuan uses a constant single-class harm fallback. Configuration and frozen
Hunyuan settings appear in Appendix~\ref{app:predictor-config}. \\
Candidate rule & Choose the highest-benefit action from COMPRESS/DROP,
then require $r_\theta\le\lambda$ and $\widehat b_\psi\ge\tau$.
Otherwise return KEEP, without trying the other edit. \\
Risk certification & Development-fixed candidate family and order;
independent calibration task bundles; $\delta_H=\delta_C=0.20$,
$\alpha=0.05$; one-sided binomial bounds at $\alpha/2=0.025$ per risk.
Stop at the first failed candidate. \\
Reported efficiency & Continuation prompt-token saving, not demonstrated
net runtime or monetary saving. No summary LLM call is used; text-processing,
feature-extraction, and selector runtime overhead are not separately itemized. \\
\bottomrule
\end{tabularx}
\end{table}

\subsection{Policy evaluation at one checkpoint}
Given a restored checkpoint $x=(s,B)$, a frozen feature map and predictors,
and selected thresholds $(\lambda,\tau)$:
\begin{enumerate}[leftmargin=*,itemsep=2pt]
\item Compute $z=\phi(B,s)$ using only the available prefix and training-fitted
preprocessing. No continuation outcome is an input to this step.
\item Evaluate both edits' normalized-benefit scores and select $a_b$ using
COMPRESS-first tie rule. Evaluate the harm score for that selected edit.
\item If both gates in Equation~\eqref{eq:models-policy} pass, apply $a_b$ to
$B$ using the transformations in Appendix~\ref{app:block-transform}, including
DROP's duplicate-text cleanup, while preserving the environment. Otherwise
return KEEP; do not try the lower-ranked edit.
\item Continue the frozen backbone from the checkpoint. This invocation
contains one selected history intervention, not the full sequence of
checkpoint decisions used in the offline comparison.
\end{enumerate}

\subsection{Independent policy certification}
Development fixes the feature preprocessing, fitted models, thresholds,
ordered candidates, and tie rules before calibration outcomes are read.
Starting from $\widehat\pi=\pi_0\equiv\Kact$, process candidates in that order.
For each candidate, evaluate all retained checkpoints of each calibration
task from the frozen action matrix, form $\ell_H$ and $\ell_C$ including
zero-loss empty bundles under the registered-task convention, and compute
$k_H,k_C,U_H,U_C$. Here $\ell_C$ sums normalized benefits. If either bound
fails its budget, stop immediately. Otherwise update $\widehat\pi$ only when
the candidate makes at least one calibration edit, and continue. Return the
last stored policy, retaining KEEP when there is no such candidate. Do not reorder by calibration savings, skip a failure, or
choose a threshold from test outcomes.

These procedures use the same frozen inputs throughout calibration and
test. Appendix~\ref{app:implementation-status} gives their concrete feature,
transformation, model, and candidate-construction rules.

\FloatBarrier
\section{Hunyuan Evaluation Protocol and Detailed Results}
\label{app:hunyuan-results}

The Hunyuan experiment uses Tencent Hunyuan-A13B-Instruct with the official
FP8 checkpoint. Each retained checkpoint has three matched seeds per action.
The seven selectors reuse the same counterfactual outcomes rather than
resampling their KEEP references. The reported results include registered-task
counts, retained-checkpoint counts, and the exact selected calibration bounds.

\subsection{Registered tasks versus retained-state tasks}

There are 19 registered Hunyuan test tasks, of which 15 have retained states.
Table~\ref{tab:main-crossfamily} uses the registered-task denominator (19),
consistent with the calibration convention. The original retained-task
comparison is preserved here: Always-COMPRESS has
$3/15=20.00\%$ task harm, whereas counting all registered tasks gives
$3/19=15.79\%$. Always-DROP similarly gives $1/15=6.67\%$ versus
$1/19=5.26\%$. Registered tasks with no retained checkpoints contribute zero
selected-edit losses; they do not supply additional continuation evidence.
Table~\ref{tab:hunyuan-denominators} presents both estimands. Four of 19
test tasks (21.05\%) have empty bundles. The development count is ten of
24 (41.67\%); the calibration empty-bundle count is not reported.

For a fixed policy, let $A=\{|X_E|>0\}$ with $\Pr(A)>0$. Under a
registered-task sampling protocol, losses are zero on $A^c$, so
\begin{equation}
R_r^{\mathrm{reg}}(\pi)
=\Pr(A)\,R_r^{\mathrm{ret}}(\pi),\qquad
R_r^{\mathrm{ret}}(\pi)=\mathbb E[\ell_r(E,\pi)\mid A],
\quad r\in\{H,C\}.
\label{eq:population-conditioning}
\end{equation}
An upper bound on $R_r^{\mathrm{reg}}$ is not the same numerical bound on
$R_r^{\mathrm{ret}}$. Applying Theorem~\ref{thm:certificate} requires that
calibration and future bundles follow the same sampling convention.

\begin{table}[ht]
\centering
\caption{\textbf{Hunyuan denominator conventions.} The second percentage
uses all 19 registered test tasks with zero observed intervention failures for
those without retained checkpoints. It is a different estimand from the
15-task conditional comparison, not additional evaluation evidence.}
\label{tab:hunyuan-denominators}
\small
\setlength{\tabcolsep}{3.5pt}
\begin{tabular*}{\linewidth}{@{\extracolsep{\fill}}lrrrr@{}}
\toprule
Method & \shortstack{Harmful\\tasks} & /15 retained & /19 registered & \shortstack{Harmful/\\switches} \\
\midrule
KEEP & 0 & 0.00\% & 0.00\% & -- \\
Always-COMPRESS & 3 & 20.00\% & 15.79\% & 3/37 \\
Always-DROP & 1 & 6.67\% & 5.26\% & 1/37 \\
RECENCY & 0 & 0.00\% & 0.00\% & 0/4 \\
SEMANTIC & 1 & 6.67\% & 5.26\% & 1/20 \\
LEARNED-NOCERT & 0 & 0.00\% & 0.00\% & 0/16 \\
RECAP & 0 & 0.00\% & 0.00\% & 0/16 \\
\bottomrule
\end{tabular*}
\end{table}

Development has 24 registered tasks; 14 contribute the 32 states used to
select Hunyuan's heuristic thresholds. The calibration task ledger contains
30 registered tasks: 15 contribute 41 retained checkpoints and 15 have empty
bundles. The calibration bounds retain the 30-task denominator and cannot
be reinterpreted as conditional retained-task bounds. These counts come from
the calibration task ledger, not by extrapolation from development or test.

\paragraph{Pointwise precision of zero observed task harm.}
For a prespecified fixed policy and iid task-bundle Bernoulli losses, zero
events among $m$ tasks give the one-sided 95\% upper limit
$1-0.05^{1/m}$. Thus the existing RECAP test counts imply approximately
14.6\% for $0/19$ registered tasks and 18.1\% for $0/15$ retained tasks.
These are deterministic pointwise calculations from the reported counts,
not new rollouts, task-bootstrap intervals, or the $\alpha/2=0.025$
calibration bounds. Their populations differ, and they do not provide a
simultaneous statement over methods or backbones.

\paragraph{Unconditional cost arithmetic.}
The main table computes each Hunyuan $S_{\rm all}$ as 217,219.0 minus the
corresponding mean cost in Table~\ref{tab:hunyuan-ledger}, then reports
$100S_{\rm all}/217{,}219.0$. The resulting values use all 111 continuations
and include zero KEEP differences. Values derived from displayed one-decimal
means inherit that rounding. The more precise recorded RECAP saving
4,619.8288 is retained below. No unreported backbone costs are imputed.

\subsection{Seven-method cost and action ledger}
\begin{table}[ht]
\centering
\caption{\textbf{Hunyuan test ledger.} All rows contain 111 selected
continuations (37 states $\times$ 3 seeds). $N_{\rm pair}$ counts joint-success
pairs; the saving column averages over those pairs, including unchanged KEEP
comparisons. Mean prompt cost averages over all 111 continuations.}
\label{tab:hunyuan-ledger}
\small
\setlength{\tabcolsep}{3.5pt}
\begin{tabular}{lrrrr}
\toprule
Method & K/C/D & $N_{\rm pair}$ & Paired saving & Mean prompt cost \\
\midrule
KEEP & 37/0/0 & 102 & +0.0 & 217,219.0 \\
Always-COMPRESS & 0/37/0 & 95 & +8,700.6 & 205,604.8 \\
Always-DROP & 0/0/37 & 100 & +20,235.7 & 197,922.5 \\
RECENCY & 33/4/0 & 102 & -300.6 & 217,482.0 \\
SEMANTIC & 17/0/20 & 100 & +14,969.9 & 202,831.5 \\
LEARNED-NOCERT & 21/10/6 & 102 & +5,000.4 & 212,599.2 \\
\textbf{RECAP} & 21/10/6 & 102 & +5,000.4 & 212,599.2 \\
\bottomrule
\end{tabular}
\end{table}

The RECAP paired saving is
$5{,}000.4118$ over 102 jointly successful pairs, including zero differences
for KEEP decisions. Its unconditional saving is $4{,}619.8288$ prompt tokens
per continuation and $4{,}594.9730$ total tokens per continuation. Thus a
paired-success average must not be multiplied by the 16 interventions to
estimate total saving. RECAP and LEARNED-NOCERT
have equal aggregates on all listed test metrics. Equality of their individual
checkpoint actions cannot be established from aggregate counts alone.

\paragraph{Hunyuan heuristic configuration.}
SEMANTIC uses the frozen
\texttt{sentence-transformers/all-MiniLM-L6-v2} encoder, 384-dimensional
normalized embeddings, and a 512-token input limit. It drops a block when
cosine similarity is strictly below $0.4563114941$; otherwise it keeps the
block. RECENCY compresses when block age is strictly greater than zero.
Candidate quantiles are $0.25,0.5,0.75$ for both heuristics, with selected
quantiles $0.25$ and $0.5$, respectively. Threshold selection uses development
outcomes, in the recorded priority order: maximize success, minimize task
harm, maximize paired-success token saving, maximize diagnostic preference
agreement, and minimize switching. The test set does not select thresholds.
The semantic checks exclude relational features and future outcomes.

\subsection{Candidate calibration ledger}
\begin{figure}[tbp]
\centering
\begin{minipage}[t]{0.48\linewidth}
\centering
\begin{tikzpicture}
\begin{axis}[
  recap/candidate axis,width=\dimexpr\linewidth-8pt\relax,
  title={(a) Task-level upper risk bounds},
  ymin=0,ymax=0.245,ytick={0,0.05,0.10,0.15,0.20},
  ylabel={Upper risk bound},
  legend style={at={(0.04,0.04)},anchor=south west},legend columns=2
]
\addplot[RECAPPlotBlue,mark=*] coordinates {(1,0.115703308) (2,0.115703308) (3,0.115703308) (4,0.115703308) (5,0.115703308)};
\addlegendentry{$U_H$}
\addplot[RECAPPlotAmber,dashed,mark=square*] coordinates {(1,0.115703308) (2,0.172169456) (3,0.172169456) (4,0.172169456) (5,0.172169456)};
\addlegendentry{$U_C$}
\addplot[RECAPPlotGray,densely dashed,mark=none,forget plot]
 coordinates {(0.7,0.20) (5.3,0.20)};
\node[anchor=south east,font=\fontsize{8}{9}\selectfont,text=RECAPPlotGray]
 at (axis cs:5.25,0.203) {Budget $0.20$};
\end{axis}
\end{tikzpicture}
\end{minipage}\hfill
\begin{minipage}[t]{0.49\linewidth}
\centering
\begin{tikzpicture}
\begin{axis}[
  recap/axis,height=4.5cm,title={(b) Calibration coverage and benefit},
  xmin=0,xmax=28,ymin=0,ymax=3350,
  xtick={0,5,10,15,20,25},ytick={0,1000,2000,3000},
  xlabel={Edited checkpoints (\%)},ylabel={Tokens saved / checkpoint}
]
\addplot[only marks,RECAPPlotGray,mark=*,mark size=2.6pt]
 coordinates {(4.87804878,371.056911) (12.195122,2786.50407) (21.9512195,2372.45528) (19.5121951,2270.91057)};
\addplot[only marks,RECAPPlotTeal,mark=star,mark size=4pt]
 coordinates {(24.3902439,2379.77236)};
\node[anchor=south,font=\fontsize{8}{9}\selectfont] at (axis cs:4.87804878,430) {1};
\node[anchor=south,font=\fontsize{8}{9}\selectfont] at (axis cs:12.19512195,2850) {2};
\node[anchor=south east,font=\fontsize{8}{9}\selectfont] at (axis cs:21.95121951,2450) {3};
\node[anchor=north east,font=\fontsize{8}{9}\selectfont] at (axis cs:19.51219512,2200) {4};
\node[anchor=south east,font=\fontsize{8}{9}\selectfont\bfseries,text=RECAPPlotTeal]
 at (axis cs:27,2900) {5: selected};
\draw[RECAPPlotTeal,line width=0.6pt]
 (axis cs:25.5,2820) -- (axis cs:24.39024390,2450);
\end{axis}
\end{tikzpicture}
\end{minipage}
\caption{\textbf{Hunyuan calibration: admissibility is not maximum sample saving.}
The five distinct candidates are shown in their recorded order, with
$\tau=(0.10,0.05,0.01,0.02,0.00)$ and common $\lambda=0.0076923077$;
nine identical quantile-labeled rows per candidate are collapsed only for display.
Left: one-sided bounds from 30 registered calibration tasks. All candidates
pass both $0.20$ tests; these are not simultaneous confidence bands for all candidates.
Right: coverage and raw mean saving over 41 retained checkpoints, including
KEEP decisions. Candidate 5 is the final accepted policy, not the largest-saving
candidate. These are calibration summaries, not held-out tradeoff estimates.}
\label{fig:hunyuan-calibration-path}
\end{figure}
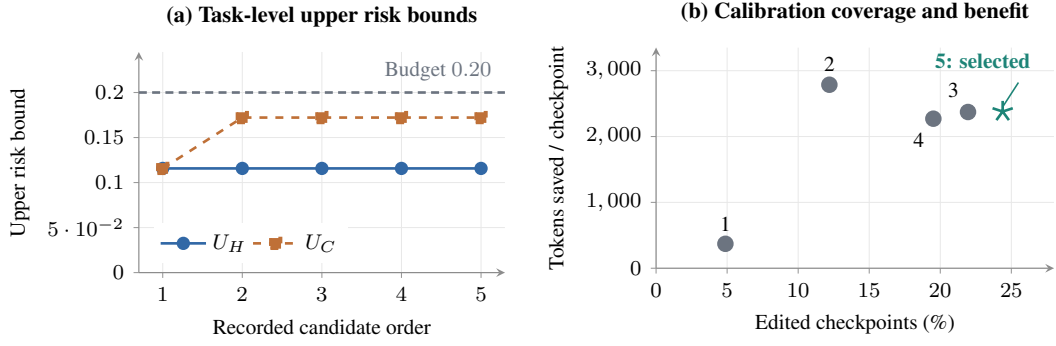

Hunyuan has 24 registered development tasks with 32 retained states, 30
calibration tasks with 41 states, and 19 registered test tasks, of which 15
contribute 37 states. The corresponding action matrices contain 288, 369,
and 333 continuations. Its selected $(\lambda,\tau)=(0.0076923077,0)$ yields
$k_H=0$, $k_C=1$, $U_H=0.1157033$, and $U_C=0.1721695$ and selects ten
calibration edits (K/C/D $=31/7/3$). The recorded grid has five distinct
threshold pairs repeated under nine quantile labels; the 45 rows are not 45
distinct policies. Figure~\ref{fig:hunyuan-calibration-path} displays these
five distinct operating points without changing their recorded order.
The last passing candidate has the highest observed switching frequency,
but not the largest calibration-sample token saving. Full counts appear in
Appendix~\ref{app:hunyuan-results}.

\begin{table}[ht]
\centering
\caption{\textbf{Distinct Hunyuan calibration operating points in fixed
order.} Each numerical row occurs under nine quantile labels in the calibration record.
All rows passed both pre-specified $0.20$ risk tests. Savings here are the
CSV's unconditional mean raw prompt-token saving per retained state.}
\label{tab:hunyuan-cal-grid}
\small
\setlength{\tabcolsep}{5pt}
\begin{tabular}{rrrrrrr}
\toprule
$\tau$ & $k_H$ & $k_C$ & $U_H$ & $U_C$ & Switches & Saving/state \\
\midrule
0.10 & 0 & 0 & 0.1157 & 0.1157 & 2/41 & 371.06 \\
0.05 & 0 & 1 & 0.1157 & 0.1722 & 5/41 & 2,786.50 \\
0.01 & 0 & 1 & 0.1157 & 0.1722 & 9/41 & 2,372.46 \\
0.02 & 0 & 1 & 0.1157 & 0.1722 & 8/41 & 2,270.91 \\
0.00 & 0 & 1 & 0.1157 & 0.1722 & 10/41 & 2,379.77 \\
\bottomrule
\end{tabular}
\end{table}

The 45 CSV rows comprise nine repeated quantile labels for each of the five
operating points in Table~\ref{tab:hunyuan-cal-grid}. All have
$\lambda=0.007692307692307694$ and all have \texttt{passed=1}. They appear in
the recorded order $\tau=0.10,0.05,0.01,0.02,0.00$. Hence neither the recorded
order nor its coverage should be described as strictly monotone. The final
nine rows share $\tau=0$, ten switches, K/C/D $=31/7/3$, zero harmful-task
flags, one cost-failure flag, and 97,570.67 aggregate prompt tokens saved
across independent calibration continuations. The selected
policy fixes $\tau=0$. All five threshold pairs pass, so the fixed-sequence
rule returns the final accepted pair without changing the recorded order.

The calibration counts reproduce the stated binomial bounds. Task-level
cost-failure labels sum normalized checkpoint benefits, using the same task
membership as Equation~\eqref{eq:task-risks}; raw-token sums are separate columns.
The state-decision and task-loss ledgers reproduce the selected candidate's
zero harmful-task and one cost-failure outcomes.

\subsection{Cross-backbone transfer diagnostic}
A Qwen-derived policy applied zero-shot to Hunyuan edits one of 37 states
(2.70\%) and saves 5.9 paired-success tokens. Refitting and recalibrating for
Hunyuan raise these values to 43.24\% and 5,000.4 without changing observed
success in this matrix. This recorded comparison does not isolate refitting
from recalibration and is distinct from the within-backbone main comparison.

\FloatBarrier
\section{Certification Details and Proof}
\label{app:certificate}

For $k$ failures among $m$ independent task bundles, the one-sided
Clopper--Pearson upper limit at tail probability $\gamma$ is
\begin{equation}
U_{\rm CP}(k,m;\gamma)=
\begin{cases}
F^{-1}_{\operatorname{Beta}(k+1,m-k)}(1-\gamma),&k<m,\\
1,&k=m.
\end{cases}
\label{eq:cp-appendix}
\end{equation}
With $m=30$ and $\gamma=0.025$, $k=0$ gives $0.1157033082$, and $k=1$ gives
$0.1721694556$, matching the Hunyuan calibration record. These are calibration
bounds for the recorded task-level losses, not confidence limits on the
0/16 test harmful-switch proportion.

\paragraph{Full fixed-sequence argument.}
Condition on development data so that the ordered candidates
$\pi_1,\ldots,\pi_M$ and the sampling protocol are fixed. For any fixed
candidate and risk coordinate, iid task bundles produce iid Bernoulli losses;
Equation~\eqref{eq:cp-appendix} has the stated one-sided coverage. If all
candidates satisfy both population constraints, every possible output is
admissible. Otherwise, let $j^\dagger$ be the first inadmissible candidate in
the fixed order. Every earlier candidate is admissible. Because testing stops
at the first failure, selecting any inadmissible output requires accepting
$\pi_{j^\dagger}$. At least one coordinate $r\in\{H,C\}$ has
$R_r(\pi_{j^\dagger})>\delta_r$. Acceptance requires
$U_r(\pi_{j^\dagger})\le\delta_r<R_r(\pi_{j^\dagger})$, an error controlled
by the exact one-sided test. A conservative union bound over the two risk
coordinates gives probability at most $\alpha$. KEEP has identically zero
losses. This proves Theorem~\ref{thm:certificate}.

\paragraph{Assumptions and interpretation.}
The proof uses independence and a common distribution across task bundles;
exchangeability alone does not establish the binomial model. It also assumes
a fixed sample-size protocol, not optional sampling until certification
succeeds. The statement allows arbitrary dependence within a task and does
not require correct predictive-model probabilities. It does not provide
simultaneous confidence intervals for all post-selected candidates, a global
coverage optimum, or a joint certificate over all separately tested backbones.
Finally, the event being bounded uses measured finite-seed outcome differences.
A guarantee for latent task-success probabilities or sequential curation
requires additional measurement or evaluation arguments.

\subsection{Interpretation of the coverage objective}
Equation~\eqref{eq:selection-objective} describes the motivating population
tradeoff. The implemented fixed-sequence procedure searches a candidate
order fixed during development and returns its last accepted member with
nonzero calibration edits, or KEEP. It does not require nestedness of the
two-threshold family.

\subsection{A text-only decision observation}
\begin{proposition}[Context alone need not determine the decision]If the optimal-action sets for the same block $B$ at two checkpoints are
disjoint, no deterministic rule using only $B$ is optimal at both checkpoints.
\end{proposition}
\begin{proof}
Such a rule must return the same action for identical $B$, whereas no action
belongs to both optimal-action sets.
\end{proof}
The proposition motivates conditioning on execution, without asserting that
any particular feature representation is sufficient.

\FloatBarrier
\section{Counterfactual Evaluation and Feature Details}
\label{app:replay}

\paragraph{Replay and intervention boundaries.}
All selector baselines read the same K/C/D outcome matrix. A KEEP decision
therefore reuses the identical reference outcome, avoiding differences caused
solely by independent model sampling. Prefix replay checks tool names,
arguments, observations, environment-state hash, and block identity before
branching. COMPRESS and DROP alter the selected context block, not the true
environment or the historical actions. Subsequent continuations use the fixed
backbone configuration and seed protocol.

\subsection{Configuration of the fixed-block case}
\label{app:same-block-case}
Figure~\ref{fig:same-block-longitudinal} uses the course-selection trajectory
\texttt{course\_r5\_c5\_h1\_b0} with the recorded
\texttt{openai/Qwen/Qwen3-8B} backend. Its seven checkpoints are
$15,17,19,21,23,25,27$, with sampling seeds $0,1,2,3,4$ for each action.
The recorded decoding settings are temperature $0.1$, top-$p=0.95$,
maximum output tokens $1{,}024$, and \texttt{enable\_thinking=false}.
These settings describe this illustrative run only, not all backbone evaluations.

The fixed block contains six failed slot checks at prefix steps
$5,7,9,11,13,15$ for slot $(2,3)$. It contains the corresponding six assistant
tool-call messages and six tool responses. Its compressed replacement is:
\begin{quote}
\small
Previously tested candidates for slot (2,3): [\texttt{\char34 C569\char34},
\texttt{\char34 C507\char34}, \texttt{\char34 C283\char34},
\texttt{\char34 C286\char34}, \texttt{\char34 C769\char34},
\texttt{\char34 C364\char34}]. These candidates were invalid under the states
in which they were tested.
\end{quote}
The block and each treatment's transformed-block hash are identical across
checkpoints; each checkpoint's environment hash matches across treatments.
The raw record reports successful prefix tool-output matching.
This establishes which summary was evaluated, not the general algorithm
used to generate summaries in every experiment.

For this case, \texttt{success} is corroborated by the recorded terminal
reward/status. Recovery steps locate the first valid check of the target slot;
they need not equal the number of calls until task termination.
The continuation limit is $50-t$ at checkpoint $t$. At step 15, COMPRESS and
DROP exhaust their 35-step limits without recovery: the aggregate value 36
is the failure penalty (limit plus one), not a successful recovery time.
The cost plot uses actual \texttt{total\_prompt\_tokens} for every run,
including failed runs, rather than penalized recovery steps.

For plotted cost, let $\overline C_t^K$ be the mean KEEP prompt cost over the
five seeds at checkpoint $t$. The displayed point for action $a$ is
$\frac{1}{5}\sum_{j=0}^{4}\Cost_j(x_t,a)/\overline C_t^K$; its whisker endpoints
are the minimum and maximum of those five normalized costs. KEEP costs are
identical across seeds at each checkpoint in this record. A collapsed range
therefore means identical observed values, not zero population uncertainty.
All 21 aggregate rows in \texttt{matrix.csv} agree with the corresponding
raw success, prompt-token, repetition, and penalized recovery statistics.
Neither the 105 runs nor the seven checkpoints are 105 or seven independent
tasks. No task-level bootstrap is reported for this single-task case.

\paragraph{Longitudinal outcomes.}
Figure~\ref{fig:same-block-longitudinal} extends the two checkpoints in
Figure~\ref{fig:state_dependent} to all seven recorded checkpoints of one
Qwen3-8B course-selection trajectory. The six failed checks and their
compressed replacement remain identical throughout. At step 15, both edits
fail in all five seeds despite shortening the first prompt. At step 27,
COMPRESS preserves observed success and reduces mean continuation prompt
cost from 23,754 to 21,846 tokens (8.03\%), whereas DROP costs 3.45 times KEEP.
The relation is not monotone: at step 21, DROP saves 154 tokens but takes two
additional recovery steps. Thus the token-based oracle in
Equation~\eqref{eq:oracle} selects DROP there, whereas the recovery-first
preference diagnostic selects KEEP. The figure reports measurements rather
than treating those objectives as interchangeable. This is a within-task
case study, not an estimate of prevalence across tasks; its configuration
and summary text are given above.

\paragraph{Worked empirical decision.}
At checkpoint 27, KEEP and COMPRESS succeed in all five recorded seeds,
so $\Delta_Y(x,\Cact)=0$ and $h(x,\Cact)=0$. Mean prompt costs 23,754 and
21,846 give $\Delta_C(x,\Cact)=1,908>0$. DROP has higher cost than KEEP,
so Equation~\eqref{eq:oracle} selects COMPRESS. This calculation illustrates
the measured oracle; it is separate from the learned score-based decision
in Equation~\eqref{eq:models-policy}.

\subsection{Feature construction and implementation details}
\label{app:implementation-status}
This section specifies the relational feature map, context transformations,
training targets, predictor construction, and ordered candidate tests used by
the implementation. The generic dataset builder is distinct from the fixed
six-failure example in Section~\ref{app:same-block-case}: in the generic
builder the selected block may change between checkpoints. The serialized
Hunyuan predictors and calibration records provide a concrete, hash-linked
instance of the shared training and certification pipeline.

\subsubsection{The ordered 72-feature representation}
\label{app:feature-schema}
\paragraph{Prefix objects and normalization.}
The extractor operates on the tool-event prefix through checkpoint step $t$.
Each event carries the tool name, arguments, and recorded tool output. Its
entity, slot, constraint, and tool sets are denoted by $D_e,D_s,D_c,D_u$.
Entity IDs come from the argument fields \texttt{id}, \texttt{candidate\_id},
\texttt{item\_id} and their list-valued plural forms. For candidate-query
outputs, IDs in \texttt{matches} and \texttt{items} are also included.
Slots come from integer-convertible \texttt{row}/\texttt{col} pairs in arguments
or output data. For a failure, the extractor unions the failure event's sets
with those of the immediately preceding state-update trigger, when present;
the inferred candidate ID is added to its entity set. IDs are used for set
comparisons but are not themselves numerical model inputs.

Tool names are mapped, in order, to \texttt{query\_candidate} (names containing
both \texttt{query\_} and \texttt{candidate}), \texttt{set\_slot},
\texttt{validate\_slot} (\texttt{check\_} and \texttt{slot\_constraints}),
\texttt{validate\_global} (\texttt{check\_} and \texttt{global\_constraints}),
\texttt{get\_state} (\texttt{get\_current\_grid\_state}),
\texttt{get\_attributes} (\texttt{item\_attributes}),
\texttt{get\_info} (\texttt{item\_info}), \texttt{done}, or \texttt{other}.
These map respectively to the action categories query, update, validate,
observe, finish, and other, with the two validation tools sharing validate
and the three getter tools sharing observe. The current subgoal is slot if
the last event contains a slot; otherwise global for global validation;
otherwise entity if an entity is present; otherwise none. ``Current'' means
the last executed prefix event, not a predicted next tool call.

Constraint tokens are derived from an argument's \texttt{field} and from the
output \texttt{reason}, falling back to the output \texttt{messages}; validation
tools additionally supply their slot/global constraint category. The first
matching lowercase substring in the following ordered groups determines the
category: price/cost/budget $\mapsto$ cost; credits/quantity/count $\mapsto$
quantity; workload/weight $\mapsto$ load; duration/time/date/day $\mapsto$ time;
difficulty/rating/score $\mapsto$ quality; teacher/brand $\mapsto$ identity;
category $\mapsto$ category; location/city/country $\mapsto$ location;
experience/skill $\mapsto$ qualification; calories/protein/fat $\mapsto$
nutrition. If none matches, text containing both slot and constraint maps
to slot-constraint, then text containing global and constraint maps to
global-constraint; the remaining category is other-constraint.

Let $m_B$ be the number of selected failure events, and define
\[
 A_g=\bigcup_{f\in B}D_g(f),\qquad
 F_g=\bigcup_{e\in\mathrm{last4}(t)}D_g(e),
 \qquad g\in\{e,s,c,u\}.
\]
The frontier is the last four available tool events (or the whole prefix if
shorter), not a separately learned planning frontier. Let $t_0$ and $t_B$
be the earliest and latest block-failure steps. Let $U^+$ and $U^-$ be prefix
\texttt{set\_slot} events with and without output status exactly
\texttt{success}; the suffix $>t_B$ restricts them to steps after $t_B$.
A successful update counter does not test whether the assigned value changed.

\paragraph{Failure status.}
For each block failure, only later events that remain inside the checkpoint
prefix are inspected. Resolution is recorded after a valid slot check on an
overlapping slot, or after a valid global check for a failure carrying the
global-constraint token. In the remaining cases, a \texttt{tool\_error} can be
resolved by a successful event with an overlapping normalized tool set.
A retry is a later \texttt{set\_slot} on an overlapping slot that assigns the
same known candidate. Supersession is a successful \texttt{set\_slot} on an
overlapping slot with a different candidate. In particular, supersession
does not require subsequent validation success. These are operational
prefix-derived indicators, not proofs that a record has become dispensable.

\paragraph{Dependency graph.}
There is one node for each prefix tool event. A directed edge from an earlier
node $i$ to a later node $j$ is inserted when at least one of these conditions
holds: their entity, slot, or constraint sets overlap; both are updates to an
overlapping slot; $j$ is a validation and they overlap in entity/slot/constraint;
or they have the same normalized tool and overlap in entity/slot/constraint.
The corresponding edge types are \texttt{same\_entity}, \texttt{same\_slot},
\texttt{same\_constraint}, \texttt{modifies\_same\_state\_field},
\texttt{validation\_of}, and \texttt{retries}. Two updates to an overlapping
slot additionally receive \texttt{supersedes} when $j$ succeeds and assigns
a different candidate. A pair can carry several labels. Thus the retries
edge is an overlap heuristic, not an equality test on complete arguments.
Breadth-first search gives each block failure's shortest distance $d_f$ to
any last-four-event node. A source already in that frontier has distance
zero. Distances that are unreachable are excluded from the finite-distance
mean; if all are unreachable, the minimum and mean are both set to 51.
Let $M_G$ count distinct directed pairs, and let $N_h$ count incidences of
edge label $h$; edge-type fractions divide by $\max(1,\sum_hN_h)$, not $M_G$.

\paragraph{Complete schema.}
Table~\ref{tab:implementation-features} preserves the serialized order:
43 overlap/relevance features, 14 temporal/state-change features, and 15
dependency-graph features. Some coordinates are deliberately redundant in
the recorded representation; they have not been removed after training.
Empty sets produce zero counts, empty-union Jaccard is zero, and unknown
categories use other/none as specified above. Malformed required records
raise validation errors rather than being silently imputed. In particular,
there is no blanket learned missing-value imputation step.

\begingroup
\footnotesize
\setlength{\tabcolsep}{3pt}
\setlength{\LTleft}{0pt}
\setlength{\LTright}{0pt}
\renewcommand{\arraystretch}{1.12}
\begin{longtable}{@{}r>{\raggedright\arraybackslash}p{0.35\linewidth}>{\raggedright\arraybackslash}p{0.58\linewidth}@{}}
\caption{\textbf{Ordered relational feature dictionary.} All quantities use the observed prefix and selected failures. The two action indicators are appended after these 72 coordinates.}\label{tab:implementation-features}\\
\toprule
No. & Feature name & Raw computation \\
\midrule
\endfirsthead
\multicolumn{3}{l}{\textit{Table~\ref{tab:implementation-features} continued}}\\
\toprule
No. & Feature name & Raw computation \\
\midrule
\endhead
\midrule
\multicolumn{3}{r}{\textit{Continued on the next page}}\\
\endfoot
\bottomrule
\endlastfoot
\multicolumn{3}{l}{\textbf{Overlap and relevance (43)}}\\
1 & \path{entity_overlap} & $\ind\{A_e\cap F_e\ne\varnothing\}$. \\
2 & \path{slot_overlap} & $\ind\{A_s\cap F_s\ne\varnothing\}$. \\
3 & \path{constraint_overlap} & $\ind\{A_c\cap F_c\ne\varnothing\}$. \\
4 & \path{tool_overlap} & $\ind\{A_u\cap F_u\ne\varnothing\}$. \\
5 & \path{entity_jaccard} & $|A_e\cap F_e|/|A_e\cup F_e|$; zero for an empty union. \\
6 & \path{slot_jaccard} & $|A_s\cap F_s|/|A_s\cup F_s|$; zero for an empty union. \\
7 & \path{constraint_jaccard} & $|A_c\cap F_c|/|A_c\cup F_c|$; zero for an empty union. \\
8 & \path{tool_jaccard} & $|A_u\cap F_u|/|A_u\cup F_u|$; zero for an empty union. \\
9 & \path{failed_candidate_still_active} & One iff a block candidate is in the latest successful \texttt{set\_slot} assignment map at the checkpoint. \\
10 & \path{same_subgoal_as_failure} & $\ind\{A_s\cap F_s\ne\varnothing\ \lor\ A_c\cap F_c\ne\varnothing\}$. \\
11 & \path{same_entity_as_failure} & $\ind\{A_e\cap F_e\ne\varnothing\}$; intentionally identical to \texttt{entity\_overlap}. \\
12 & \path{block_relevance_ratio} & Fraction of block failures whose entity, slot, constraint, or tool set overlaps the corresponding frontier set. \\
13 & \path{affected_entity_count} & $|A_e|$. \\
14 & \path{affected_slot_count} & $|A_s|$. \\
15 & \path{affected_constraint_count} & $|A_c|$. \\
16 & \path{affected_tool_count} & $|A_u|$. \\
17 & \path{frontier_entity_count} & $|F_e|$. \\
18 & \path{frontier_slot_count} & $|F_s|$. \\
19 & \path{frontier_constraint_count} & $|F_c|$. \\
20 & \path{frontier_tool_count} & $|F_u|$. \\
21 & \path{context_block_size} & $m_B$, the number of selected failure events, not the number of tokens or messages. \\
22 & \path{current_action_type__query} & One iff the last prefix event has normalized action type \texttt{query}. \\
23 & \path{current_action_type__update} & One iff the last prefix event has normalized action type \texttt{update}. \\
24 & \path{current_action_type__validate} & One iff the last prefix event has normalized action type \texttt{validate}. \\
25 & \path{current_action_type__observe} & One iff the last prefix event has normalized action type \texttt{observe}. \\
26 & \path{current_action_type__finish} & One iff the last prefix event has normalized action type \texttt{finish}. \\
27 & \path{current_action_type__other} & One iff the last prefix event has normalized action type \texttt{other}. \\
28 & \path{current_tool__query_candidate} & One iff the last prefix event has normalized tool \path{query_candidate}. \\
29 & \path{current_tool__set_slot} & One iff the last prefix event has normalized tool \path{set_slot}. \\
30 & \path{current_tool__validate_slot} & One iff the last prefix event has normalized tool \path{validate_slot}. \\
31 & \path{current_tool__validate_global} & One iff the last prefix event has normalized tool \path{validate_global}. \\
32 & \path{current_tool__get_state} & One iff the last prefix event has normalized tool \path{get_state}. \\
33 & \path{current_tool__get_attributes} & One iff the last prefix event has normalized tool \path{get_attributes}. \\
34 & \path{current_tool__get_info} & One iff the last prefix event has normalized tool \path{get_info}. \\
35 & \path{current_tool__done} & One iff the last prefix event has normalized tool \path{done}. \\
36 & \path{current_tool__other} & One iff the last prefix event has normalized tool \path{other}. \\
37 & \path{current_subgoal__slot} & One iff the last prefix event has subgoal category \texttt{slot}, assigned in the priority order defined in the text. \\
38 & \path{current_subgoal__global} & One iff the last prefix event has subgoal category \texttt{global}, assigned in the priority order defined in the text. \\
39 & \path{current_subgoal__entity} & One iff the last prefix event has subgoal category \texttt{entity}, assigned in the priority order defined in the text. \\
40 & \path{current_subgoal__none} & One iff the last prefix event has subgoal category \texttt{none}, assigned in the priority order defined in the text. \\
41 & \path{failure_type_validation_ratio} & Fraction of block events classified as \texttt{validation\_failure}. \\
42 & \path{failure_type_tool_error_ratio} & Fraction of block events classified as \texttt{tool\_error}. \\
43 & \path{failure_type_other_ratio} & Fraction in neither of the preceding two classes, including \texttt{empty\_result}. \\
\midrule
\multicolumn{3}{l}{\textbf{Temporal and state-change relations (14)}}\\
44 & \path{failed_candidate_retried_since_block} & $\ind\{\exists f\in B:\mathrm{retried}(f)\}$, using events after each failure but not after $t$. \\
45 & \path{failure_already_resolved} & $\ind\{\exists f\in B:\mathrm{resolved}(f)\}$. \\
46 & \path{failure_superseded_by_success} & $\ind\{\exists f\in B:\mathrm{superseded}(f)\}$. \\
47 & \path{state_changed_since_failure} & $\ind\{|U^+_{>t_B}|>0\}$. This counts successful updates, not a comparison of state hashes. \\
48 & \path{num_successful_updates_since_failure} & $|U^+_{>t_B}|$, successful \texttt{set\_slot} calls after the latest block failure. \\
49 & \path{num_failed_updates_since_failure} & $|U^-_{>t_B}|$, unsuccessful \texttt{set\_slot} calls after the latest block failure. \\
50 & \path{distance_in_tool_steps} & $t-t_B$. \\
51 & \path{distance_in_state_changes} & $|U^+_{>t_B}|$; identical to \texttt{num\_successful\_updates\_since\_failure}. \\
52 & \path{resolved_failure_ratio} & $m_B^{-1}\sum_{f\in B}\ind\{\mathrm{resolved}(f)\}$. \\
53 & \path{superseded_failure_ratio} & $m_B^{-1}\sum_{f\in B}\ind\{\mathrm{superseded}(f)\}$. \\
54 & \path{current_partial_solution_changes} & $|U^+|$, the total number of successful \texttt{set\_slot} calls in the prefix. \\
55 & \path{recent_partial_solution_changes} & Number of successful \texttt{set\_slot} calls among the last four prefix tool events. \\
56 & \path{distance_from_last_successful_state_change} & $t-\max_{e\in U^+}\mathrm{step}(e)$; use $t$ if $U^+$ is empty. \\
57 & \path{failures_since_block_start} & Number of all detected failures with $t_0\le\mathrm{step}(f)\le t$, not just the failures retained in $B$. \\
\midrule
\multicolumn{3}{l}{\textbf{Dependency graph (15)}}\\
58 & \path{min_graph_distance} & Minimum finite distance $d_f$ from a block failure to the last-four-event frontier; use 51 if none is reachable. \\
59 & \path{mean_graph_distance} & Mean over finite $d_f$ only; use 51 if none is reachable. \\
60 & \path{reachable_failure_fraction} & $m_B^{-1}\sum_{f\in B}\ind\{d_f<\infty\}$. \\
61 & \path{active_path_failure_fraction} & Fraction of block failures that reach the frontier and are neither resolved nor superseded. \\
62 & \path{superseded_branch_fraction} & Fraction of block failures marked superseded; identical to \texttt{superseded\_failure\_ratio}. \\
63 & \path{graph_node_count} & $n$, the number of prefix tool events. \\
64 & \path{graph_edge_count} & $M_G$, the number of distinct forward event pairs with at least one edge type. \\
65 & \path{graph_density} & $M_G/[n(n-1)/2]$; zero when the denominator is zero. \\
66 & \path{graph_edge_fraction__same_entity} & $N_{\mathrm{same\_entity}}/\max(1,\sum_h N_h)$, counting edge-type incidences rather than distinct pairs. \\
67 & \path{graph_edge_fraction__same_slot} & $N_{\mathrm{same\_slot}}/\max(1,\sum_h N_h)$, counting edge-type incidences rather than distinct pairs. \\
68 & \path{graph_edge_fraction__same_constraint} & $N_{\mathrm{same\_constraint}}/\max(1,\sum_h N_h)$, counting edge-type incidences rather than distinct pairs. \\
69 & \path{graph_edge_fraction__modifies_same_state_field} & $N_{\mathrm{modifies\_same\_state\_field}}/\max(1,\sum_h N_h)$, counting edge-type incidences rather than distinct pairs. \\
70 & \path{graph_edge_fraction__validation_of} & $N_{\mathrm{validation\_of}}/\max(1,\sum_h N_h)$, counting edge-type incidences rather than distinct pairs. \\
71 & \path{graph_edge_fraction__retries} & $N_{\mathrm{retries}}/\max(1,\sum_h N_h)$, counting edge-type incidences rather than distinct pairs. \\
72 & \path{graph_edge_fraction__supersedes} & $N_{\mathrm{supersedes}}/\max(1,\sum_h N_h)$, counting edge-type incidences rather than distinct pairs. \\
\end{longtable}
\endgroup

\paragraph{Scaling and action augmentation.}
The training routine creates one row for each checkpoint--edit pair, with
edits ordered as (COMPRESS, DROP). It fits the mean $\mu_k$ and population
standard deviation $\sigma_k$ of each of the 72 raw features on the development
rows only, and uses
\begin{equation}
 z_k=\frac{v_k-\mu_k}{s_k},\qquad
 s_k=\begin{cases}1,&\sigma_k<10^{-8},\\\sigma_k,&\text{otherwise},\end{cases}
 \qquad
 \mathbf x(z,a)=\bigl[z;\ind\{a=\Cact\};\ind\{a=\Dact\}\bigr].
 \label{eq:impl-feature-scaling}
\end{equation}
Both predictors therefore receive 74 numerical inputs, not 72. State features
are duplicated once per edit when fitting the scaler, and the two action
indicators are not standardized. The fit-time scaler computes moments and
transforms from float32 arrays; its saved moments are converted to float64.
The calibration/test path applies the saved moments to float64 arrays without
refitting. This precision detail is part of the serialized implementation.

\subsubsection{Failure blocks, checkpoint sampling, and summaries}
\label{app:block-transform}
\paragraph{Failure detection and offline checkpoint selection.}
Tool calls and outputs are paired by call ID, with positional fallback when
an ID match is unavailable. The detector classifies a tool error for status
error/failed/failure/invalid; a validation failure for \texttt{is\_valid=false};
an empty result for an empty \texttt{matches} list; and otherwise a tool error
when a non-success-prefixed human-readable message contains a whole-word
error, failed, failure, or invalid. This order determines the class. A
preceding \texttt{set\_slot}, \texttt{update\_slot}, or \texttt{clear\_slot}
may supply the triggering action and candidate ID. Baseline screening
requires a successful trajectory with at least two detected natural failures.

The generic v2 builder considers post-failure events, annotated recovery
events, state-update calls, and valid constraint checks from the second
failure onward and before step 50. Recovery annotations use the first later
valid check with the same tool name for validation failures; a changed-argument
query with nonempty matches for empty results; and a changed-argument same-tool
call that is no longer classified as a failure for other errors. The
validation recovery annotation does not additionally require a matching slot.
Candidate priority is the sum of 8 for post-failure, 5 for post-recovery,
3 for a successful constraint check, and 2 for a state-update call. If the
candidate count exceeds the cap, one candidate is first selected in each
nonempty early/middle/late third of the completed baseline trajectory,
maximizing priority and then minimizing distance to the respective fractional
centers $1/6,1/2,5/6$ (earlier step breaks ties). Remaining slots maximize
minimum step distance to selected checkpoints, then priority, then prefer
earlier steps. The Hunyuan manifest sets a cap of three; the generic v2
builder defaults to eight. This is retrospective dataset sampling using the
completed baseline, not an online checkpoint scheduler. The feature vector
at each selected checkpoint still reads only that checkpoint's prefix.

\paragraph{Generic block construction.}
At each selected checkpoint, only detected failures at or before that step
are available. The block contains the latest three such events, or both
when only two are available. For every event, the builder retains its adjacent
assistant tool-call message and corresponding tool-output message, preserving
their order. It checks the assistant/tool roles and adjacency. The block is
therefore a collection of two or three message pairs that need not form one
contiguous substring; intervening messages are not automatically part of $B$.
Candidate IDs are inferred first from the triggering action and then from
the failed call's \texttt{id}, \texttt{candidate\_id}, or \texttt{item\_id}.

\paragraph{Deterministic COMPRESS.}
The implementation does not call an LLM to generate these summaries. It
constructs compact JSON records with the fields \texttt{event\_id},
\texttt{tool}, \texttt{candidate\_id}, and \texttt{result}, the last fixed to
\texttt{invalid/failure}. Records are serialized with \texttt{ensure\_ascii=False}
and compact comma/colon separators, wrapped as
\begin{quote}\small
Prior failure observations: [compact JSON event records]. These observations
apply to the states in which they were observed.
\end{quote}
COMPRESS removes the earlier selected message pairs, keeps the last selected
assistant tool-call message, and replaces only its paired tool observation
with this text. It introduces no summary-model temperature, sampling seed,
or separate summary-model call. The generic block length is bounded by the
number of selected events; no additional summary-token cap is imposed by
this function. Reusing the same block and event metadata gives the same
summary. The fixed-block experiment instead uses its separately recorded
six-candidate template; it is not described as a two/three-failure block.

\paragraph{DROP and duplicate failure text.}
DROP removes all selected call/output pairs. In the remaining assistant
messages, it additionally inspects both \texttt{content} and
\texttt{reasoning\_content}. A text fragment is removed only if it contains
a selected candidate ID as a case-insensitive substring and matches the
case-insensitive whole-word pattern
\begin{verbatim}
\b(?:invalid|failed|failure|error|violat(?:e|es|ed|ion))\b
\end{verbatim}
Fragments are split at sentence punctuation followed by whitespace/end of
text or at newlines; retained fragments are rejoined and trimmed. Thus the
implemented DROP operator also sanitizes matching duplicate failure mentions
outside the selected message pairs. It is not deletion of only a contiguous
text span. The environment and actual past execution are unchanged; other
history is unchanged except for these explicitly defined text removals.
The illustrative fixed-block branch uses the analogous scrubber without
\texttt{error} in its failure-word list.

\subsubsection{Supervision, normalization, and the cost-risk estimand}
\label{app:benefit-target}
For the three-seed generic dataset, aggregation first checks that each
checkpoint contains the full three-action by three-seed grid. For each
checkpoint and action it averages terminal success and continuation prompt
tokens over seeds. There are two training examples per checkpoint, one for
each non-KEEP edit; KEEP supplies the reference. The targets are
\begin{equation}
 \begin{aligned}
 h(x,a)&=\ind\{\Delta_Y(x,a)<0\},\\
 \overline C_K(x)&=\frac{1}{J}\sum_{j=1}^{J}\Cost_j(x,\Kact)>0,\\
 \widetilde{\Delta}_C(x,a)&=\frac{\Delta_C(x,a)}{\overline C_K(x)}.
 \end{aligned}
 \label{eq:impl-benefit-target}
\end{equation}
The forest is a regressor for $\widetilde{\Delta}_C$, not a classifier for
positive benefit. The target is a ratio of seed-averaged costs, not a mean
of seedwise relative differences. Training includes both harmful and
non-harmful edits and failed continuations: it is not restricted to jointly
successful pairs. The target is not clipped, z-scored, or log-transformed;
nonpositive KEEP denominators are rejected. A separate learnability
diagnostic labels a token-beneficial edit by
$\ind\{\Delta_Y\ge0\land\Delta_C>0\}$; that diagnostic label is not the final
forest's regression target.

The implemented task cost loss aggregates normalized advantages:
\[
 D_E(\pi)=\sum_{x\in I_E(\pi)}\widetilde{\Delta}_C(x,\pi(x)),\qquad
 \ell_C(E,\pi)=\ind\{|I_E(\pi)|>0\land D_E(\pi)\le0\}.
\]
Raw-token sums are reported separately. Positive normalization preserves the
sign of an individual checkpoint advantage but need not preserve the sign
of a sum over checkpoints with different KEEP costs. Consequently the
certificate is for nonpositive aggregate \emph{normalized} benefit, not
necessarily for nonpositive raw-token saving. This does not change the
units of the raw-token efficiency metrics in the results tables.

\subsubsection{Final predictor construction and the Hunyuan instance}
\label{app:predictor-config}
The standalone freezing script and the per-backbone driver instantiate a
logistic harm predictor with $C=1$, balanced class weights, and at most 2,000
iterations, and a normalized-benefit random forest with 100 trees,
minimum leaf size 2, and square-root feature subsampling. They fit only on
development checkpoint--edit rows. Calibration selects thresholds; it does
not refit these models. This is distinct from selecting the best model in
the representation diagnostic. The driver sets the harm seed to
$20260828+\texttt{size\_b}$ and the benefit seed ten larger; exact fitted
parameters and preprocessing are stored in each backbone's frozen artifact.
The standalone 39-task development study uses a separate freezing script
with a configurable seed (default 20260826). Each experiment retains its
own frozen predictors, preprocessing, and configuration.

\paragraph{Single-class development branch.}
With both harm classes present, the driver fits logistic regression.
For a single observed class, it sets all slopes to zero and uses the
Jeffreys-smoothed rate $p=(k+1/2)/(N+1)$ as a constant score,
with intercept $\log[p/(1-p)]$. This branch uses the observed labels only.

Hunyuan development contains 32 retained checkpoints from 14 of 24 registered
tasks, giving 64 checkpoint--edit training rows. All 64 harm labels are zero.
The serialized harm model consequently has 74 zero slopes and
\begin{equation}
 r_\theta(z,a)=\frac{0+1/2}{64+1}=\frac1{130}
 \simeq0.0076923077,\qquad
 b_0=-\log129.
 \label{eq:impl-hunyuan-constant-risk}
\end{equation}
Table~\ref{tab:implementation-hunyuan-models} lists the frozen estimator
settings. All Hunyuan risk quantiles coincide: state-dependent edit selection
therefore comes from the benefit regressor and threshold, followed by
independent task-level tests. The constant harm score is a proposal input;
task-level risk bounds are computed separately on calibration tasks.

\begin{table}[htbp]
\centering
\caption{\textbf{Frozen Hunyuan predictor configuration.} Values are stored in
its serialized artifact; the logistic optimizer is not run in the observed
single-class branch. Both estimators expect 74 inputs.}
\label{tab:implementation-hunyuan-models}
\small
\begin{tabularx}{\linewidth}{@{}p{0.25\linewidth}XX@{}}
\toprule
Setting & Harm score & Normalized-benefit predictor \\
\midrule
Estimator & Logistic-regression object; constant single-class fallback & Random-forest regressor \\
Training rows & 64; harm counts $(64,0)$ & Same 64 checkpoint--edit rows \\
Core parameters & $C=1$; balanced class weights; intercept enabled & 100 trees; squared-error splits; bootstrap enabled \\
Shape/limits & 74 zero slopes; intercept $-\log129$ & Unlimited depth; min split 2; min leaf 2; max features sqrt (8) \\
Additional settings & lbfgs; tolerance $10^{-4}$; max iterations 2000 (unused here) & No depth/leaf-node cap; no pruning ($\texttt{ccp\_alpha}=0$); no out-of-bag scoring \\
Random state & 20260841 & 20260851 \\
Stored library version & scikit-learn 1.8.0 & scikit-learn 1.8.0 \\
\bottomrule
\end{tabularx}
\end{table}
The benefit forest has 100 fitted trees and bootstrap sample size 64.
Its remaining saved settings are \texttt{min\_samples\_split=2},
\texttt{min\_weight\_fraction\_leaf=0},
\texttt{min\_impurity\_decrease=0}, \texttt{max\_samples=None},
\texttt{warm\_start=False}, and \texttt{n\_jobs=-1}; no monotonic
constraints are set. These are the settings of the frozen Hunyuan estimator.
The artifact also contains all 72 scaler means/scales, the ordered feature
names, and action order (COMPRESS, DROP). Its byte-level SHA-256 is
\path{06f19494364556fb256f2e3fd2dcfd49151370a95cd2c060ea98139f488da3d6}.
The predictor payload is identical in the preregistered and certified JSON
files; certification changes the selected policy, not these fitted objects.

\subsubsection{Candidate generation, ordering, and frozen tests}
\label{app:candidate-construction}
At each development checkpoint, both edits are scored. The higher predicted
normalized benefit determines $a_b$; ties prefer COMPRESS because the fixed
action order is (COMPRESS, DROP). Risk thresholds are computed from the harm
scores of these selected actions using NumPy's \texttt{higher} empirical
quantile at
\[
 Q=(0.01,0.02,0.05,0.10,0.15,0.20,0.30,0.40,0.50),\qquad
 \mathcal T=(0,0.01,0.02,0.05,0.10).
\]
Each quantile is paired with each benefit threshold, producing 45 indexed
candidate entries. Equal numerical thresholds are not deduplicated during
testing. For a candidate $p$, the development-only desirability quantities
are its predicted coverage $c_p$ and predicted normalized benefit per
\emph{all} development checkpoints,
\[
 b_p=\frac1{|\mathcal X_{\rm dev}|}\sum_{x\in\mathcal X_{\rm dev}}
 \ind\{r_\theta(x,a_b)\le\lambda_p,\ \widehat b_\psi(x,a_b)\ge\tau_p\}
 \widehat b_\psi(x,a_b).
\]
These are in-sample predictions of development-fitted models, not scores
computed from calibration or test outcomes. The implementation sorts
ascending by the lexicographic key
\begin{equation}
 (-c_p,-b_p,\lambda_p,-\tau_p,\texttt{candidate\_id}),
 \label{eq:impl-candidate-order}
\end{equation}
then reverses the list for fixed-sequence testing. The full ID is the last
deterministic tie-breaker. A fit stage writes the preprocessing, fitted
models, ordered candidates, and their hashes before the calibration stage
loads outcomes; the calibration stage verifies the preregistered JSON hash.

\paragraph{Recorded Hunyuan order.}
The numerical groups are tested as $\tau=0.10,0.05,0.01,0.02,0.00$, each
with quantile labels $q50,q40,q30,q20,q15,q10,q05,q02,q01$. The development
predicted edit counts for these groups are $2,10,16,16,17$ out of 32.
For $\tau=0.01$ and $0.02$, both predicted coverage and predicted aggregate
benefit tie, so the larger-$\tau$ desirability tie-breaker followed by
list reversal puts $0.01$ first. This explains the non-numerically-sorted
sequence; no post-hoc sorting by calibration saving is applied.

The selected candidate is the last member of the accepted prefix that
has at least one selected edit on the calibration sample; if none exists,
KEEP is returned. Testing still stops at the first failed candidate,
including one with zero observed edits. The implementation may materialize
all candidate summaries for auditing, but the selection function cannot
skip a failed test. Hunyuan's 45 recorded tests all pass and have nonzero
coverage, so the selected entry is \texttt{q01\_tau0.00}, with
$(\lambda,\tau)=(0.007692307692307694,0)$.

\paragraph{Calibration ledger and scope.}
There are 30 registered Hunyuan calibration tasks: 15 contribute the 41
retained checkpoints, and 15 have empty bundles. The selected policy makes
10 edits (7 COMPRESS, 3 DROP), producing zero harmful-task flags and one
nonpositive-normalized-benefit task. The one-sided limits at 0.025 are
$U_H=0.1157033082$ and $U_C=0.1721694556$. Empty bundles remain zero-loss
registered tasks, not extra continuation measurements. The 41 state decisions
sum to 97,570.67 raw prompt tokens saved; this is a sum over matched
checkpoint continuation means, not one sequential deployment episode.
The preregistered and certified file hashes match their sidecar files and the
policy identifier in the evaluation metrics, linking the development,
calibration, and evaluation records for this backbone.

\FloatBarrier
\section{Metric Definitions and Diagnostic Preference Agreement}
\label{app:diagnostic-preference}

\paragraph{Switching and success-conditioned saving.}
Let $\mathcal X$ be the retained test checkpoints and let
\[
\mathcal P_\pi=\{(x,j):x\in\mathcal X,\ 1\le j\le J,\
Y_j(x,\Kact)=Y_j(x,\pi(x))=1\}.
\]
The two metrics are
\begin{equation}
\begin{aligned}
\widehat{\operatorname{Cov}}(\pi)
&=\frac{1}{|\mathcal X|}\sum_{x\in\mathcal X}\ind\{\pi(x)\ne\Kact\},\\
S_{\rm pair}(\pi)
&=\frac{\sum_{(x,j)\in\mathcal P_\pi}
[\Cost_j(x,\Kact)-\Cost_j(x,\pi(x))]}{|\mathcal P_\pi|}.
\end{aligned}
\label{eq:evaluation-metrics}
\end{equation}
The paired mean is undefined when no jointly successful pair exists.
Unconditional saving instead averages the same cost differences over all
$J|\mathcal X|$ checkpoint--seed pairs. Both metrics retain zero differences
for KEEP decisions. Joint-success conditioning avoids crediting cost reduction
on failed pairs, but its policy-dependent denominator prevents interpreting
it as an unconditional expected saving or an average saving per edit.

\paragraph{Unconditional and relative saving.}
Let $\overline C_K=(J|\mathcal X|)^{-1}\sum_{x,j}\Cost_j(x,\Kact)$. Then
\begin{equation}
S_{\rm all}(\pi)=\frac{1}{J|\mathcal X|}\sum_{x\in\mathcal X}\sum_{j=1}^J
[\Cost_j(x,\Kact)-\Cost_j(x,\pi(x))],\qquad
S_{\rm rel}(\pi)=\frac{100S_{\rm all}(\pi)}{\overline C_K},
\label{eq:unconditional-saving}
\end{equation}
with the latter defined when $\overline C_K>0$. Unlike $S_{\rm pair}$, these
use the same continuation denominator for all selectors on one frozen
backbone matrix. They remain prompt-cost measures, not overhead-adjusted
wall-clock, monetary, or episode costs.

\paragraph{A separate preference diagnostic, not the formal oracle.}
The historical diagnostic prioritizes success, then recovery cost, repeated
failures, and token-related tie-breakers. Let $\rho(x)$ be that stored preferred
action and $\mathcal O_\rho=\{x\in\mathcal X:\rho(x)\ne\Kact\}$. We report
\begin{equation}
\operatorname{DPA}(\pi)=
\frac{1}{|\mathcal O_\rho|}\sum_{x\in\mathcal O_\rho}
\ind\{\pi(x)=\rho(x)\},
\label{eq:diagnostic-agreement}
\end{equation}
when $|\mathcal O_\rho|>0$. This is \emph{diagnostic preference agreement}
(DPA), not agreement with the token-only empirical oracle in
Equation~\eqref{eq:oracle}, and not intervention coverage.
The development counts 94/48/144 and within-task changes in 38/39 tasks use
$\rho$. A safe, token-positive edit can disagree with $\rho$, so zero DPA does
not imply that every edit is harmful or non-beneficial. The stored labels and
values are preserved in Table~\ref{tab:diagnostic-preference} rather than being
reinterpreted as results for a different objective.

\begin{table}[ht]
\centering
\caption{\textbf{Diagnostic preference agreement (\%).} Stored preference
labels are distinct from the formal empirical oracle. Values are not
recomputed and do not measure switching frequency or statistical safety.}
\label{tab:diagnostic-preference}
\small
\setlength{\tabcolsep}{5pt}
\begin{tabular*}{\linewidth}{@{\extracolsep{\fill}}lrrrrr@{}}
\toprule
Method & Qwen3-4B & Qwen3-8B & Qwen3-14B & gpt-oss & Hunyuan \\
\midrule
KEEP & 0.0 & 0.0 & 0.0 & 0.0 & 0.0 \\
Always-COMPRESS & 61.3 & 20.0 & 30.9 & 38.2 & 40.7 \\
Always-DROP & 38.7 & 80.0 & 69.1 & 61.8 & 59.3 \\
RECENCY & 12.9 & 0.0 & 4.9 & 2.6 & 0.0 \\
SEMANTIC & 3.2 & 41.2 & 30.9 & 25.0 & 29.6 \\
LEARNED-NOCERT & 0.0 & 40.0 & 21.0 & 6.6 & 25.9 \\
\textbf{RECAP} & 0.0 & 35.0 & 7.4 & 1.3 & 25.9 \\
\bottomrule
\end{tabular*}
\end{table}

\subsection{Development diagnostic and representation results}
\label{app:development-diagnostics}
\begin{table}[htbp]
\centering
\caption{\textbf{Global leave-one-task-out harm prediction.}
Reported best estimator for each representation. The semantic comparison
changes both features and classifier, so it is not a controlled feature ablation.}
\label{tab:representation}
\small
\begin{tabular}{llr}
\toprule
Representation & Reported model & ROC--AUC \\
\midrule
Structured only & Logistic regression & 0.6147 \\
Semantic + structured & MLP & 0.4997 \\
Execution-relational & Logistic regression & \textbf{0.7015} \\
\bottomrule
\end{tabular}
\end{table}
The development diagnostic contains 39 tasks, 286 states, and 2,574
matched rollouts. Its diagnostic preference ranking selects KEEP/COMPRESS/DROP
in 94/48/144 states and varies within 38 of 39 tasks. In 90 states, success
differs across treatments; in the remaining 196, success agrees but efficiency
differs. These task-level statistics do not hold the failure block fixed.
Figure~\ref{fig:state_dependent} provides a separate same-block witness, not
a population estimate of how frequently an identical block changes value.

Execution-relational features reach 0.7015 harm ROC--AUC, compared with
0.6147 for structured-only features using the same classifier family.
The semantic-plus-structured MLP reaches 0.4997, but this comparison does not
isolate representation effects from estimator choice. A relational random
forest reaches 0.7783 ROC--AUC for token-beneficial intervention prediction.
These are diagnostic results rather than identification of the final
predictor configuration or evidence of representation sufficiency.

\section{Available Domain-Level Results}
\label{app:domains}
\begin{table}[htbp]
\centering
\caption{\textbf{Per-domain continuation success (\%).}
Colors rank methods within each backbone and domain; ties share a color.
Hunyuan has no recorded per-domain breakdown and is reported only in the
aggregate comparison in Table~\ref{tab:main-crossfamily}.}
\label{tab:domain-results}
\small
\setlength{\tabcolsep}{3.1pt}
\renewcommand{\arraystretch}{1.08}
\begin{tabular}{lrrrrrr}
\toprule
Method & Course & Meal & PC-build & Shopping & Travel & Workforce \\
\midrule
\rowcolor{ModelHeader}\multicolumn{7}{l}{\textcolor{white}{\textbf{Qwen3-4B}}} \\
KEEP & \cellcolor{BestCell}100.0 & \cellcolor{BestCell}88.9 & \cellcolor{BestCell}100.0 & \cellcolor{BestCell}100.0 & \cellcolor{BestCell}98.1 & \cellcolor{BestCell}100.0 \\
Always-COMPRESS & \cellcolor{SecondCell}88.9 & 70.4 & \cellcolor{SecondCell}88.9 & \cellcolor{BestCell}100.0 & 74.1 & \cellcolor{BestCell}100.0 \\
Always-DROP & \cellcolor{ThirdCell}74.1 & 55.6 & \cellcolor{ThirdCell}66.7 & \cellcolor{BestCell}100.0 & 85.2 & \cellcolor{SecondCell}33.3 \\
RECENCY & \cellcolor{BestCell}100.0 & \cellcolor{BestCell}88.9 & \cellcolor{BestCell}100.0 & \cellcolor{BestCell}100.0 & \cellcolor{ThirdCell}94.4 & \cellcolor{BestCell}100.0 \\
SEMANTIC & \cellcolor{BestCell}100.0 & \cellcolor{ThirdCell}77.8 & \cellcolor{BestCell}100.0 & \cellcolor{BestCell}100.0 & 92.6 & \cellcolor{SecondCell}33.3 \\
LEARNED-NOCERT & \cellcolor{BestCell}100.0 & \cellcolor{SecondCell}83.3 & \cellcolor{BestCell}100.0 & \cellcolor{BestCell}100.0 & \cellcolor{SecondCell}96.3 & \cellcolor{BestCell}100.0 \\
\textbf{RECAP} & \cellcolor{BestCell}100.0 & \cellcolor{SecondCell}83.3 & \cellcolor{BestCell}100.0 & \cellcolor{BestCell}100.0 & \cellcolor{BestCell}98.1 & \cellcolor{BestCell}100.0 \\
\midrule
\rowcolor{ModelHeader}\multicolumn{7}{l}{\textcolor{white}{\textbf{Qwen3-8B}}} \\
KEEP & \cellcolor{SecondCell}88.9 & \cellcolor{BestCell}91.7 & \cellcolor{BestCell}100.0 & \cellcolor{BestCell}96.3 & \cellcolor{SecondCell}96.3 & \cellcolor{BestCell}100.0 \\
Always-COMPRESS & \cellcolor{ThirdCell}83.3 & 79.6 & \cellcolor{SecondCell}85.7 & 79.6 & 77.8 & 84.7 \\
Always-DROP & \cellcolor{BestCell}100.0 & \cellcolor{ThirdCell}82.4 & \cellcolor{SecondCell}85.7 & \cellcolor{SecondCell}94.4 & \cellcolor{ThirdCell}93.8 & \cellcolor{ThirdCell}93.1 \\
RECENCY & \cellcolor{SecondCell}88.9 & \cellcolor{BestCell}91.7 & \cellcolor{BestCell}100.0 & \cellcolor{BestCell}96.3 & \cellcolor{SecondCell}96.3 & \cellcolor{BestCell}100.0 \\
SEMANTIC & \cellcolor{BestCell}100.0 & \cellcolor{SecondCell}88.9 & \cellcolor{BestCell}100.0 & \cellcolor{SecondCell}94.4 & \cellcolor{ThirdCell}93.8 & \cellcolor{SecondCell}97.2 \\
LEARNED-NOCERT & \cellcolor{SecondCell}88.9 & \cellcolor{BestCell}91.7 & \cellcolor{BestCell}100.0 & \cellcolor{ThirdCell}90.7 & \cellcolor{BestCell}100.0 & \cellcolor{BestCell}100.0 \\
\textbf{RECAP} & \cellcolor{SecondCell}88.9 & \cellcolor{BestCell}91.7 & \cellcolor{BestCell}100.0 & \cellcolor{ThirdCell}90.7 & \cellcolor{SecondCell}96.3 & \cellcolor{BestCell}100.0 \\
\midrule
\rowcolor{ModelHeader}\multicolumn{7}{l}{\textcolor{white}{\textbf{Qwen3-14B (AWQ 4-bit)}}} \\
KEEP & \cellcolor{SecondCell}95.8 & \cellcolor{BestCell}100.0 & \cellcolor{BestCell}100.0 & \cellcolor{BestCell}100.0 & \cellcolor{SecondCell}97.0 & \cellcolor{BestCell}97.0 \\
Always-COMPRESS & 91.7 & \cellcolor{SecondCell}97.6 & \cellcolor{ThirdCell}80.0 & \cellcolor{BestCell}100.0 & 83.8 & \cellcolor{ThirdCell}91.9 \\
Always-DROP & \cellcolor{ThirdCell}93.1 & \cellcolor{ThirdCell}95.2 & \cellcolor{SecondCell}93.3 & \cellcolor{BestCell}100.0 & 83.8 & \cellcolor{SecondCell}93.9 \\
RECENCY & \cellcolor{SecondCell}95.8 & \cellcolor{BestCell}100.0 & \cellcolor{BestCell}100.0 & \cellcolor{BestCell}100.0 & \cellcolor{SecondCell}97.0 & \cellcolor{BestCell}97.0 \\
SEMANTIC & \cellcolor{ThirdCell}93.1 & \cellcolor{BestCell}100.0 & \cellcolor{BestCell}100.0 & \cellcolor{BestCell}100.0 & \cellcolor{ThirdCell}93.9 & \cellcolor{BestCell}97.0 \\
LEARNED-NOCERT & \cellcolor{BestCell}97.2 & \cellcolor{BestCell}100.0 & \cellcolor{SecondCell}93.3 & \cellcolor{BestCell}100.0 & \cellcolor{BestCell}98.0 & \cellcolor{BestCell}97.0 \\
\textbf{RECAP} & \cellcolor{SecondCell}95.8 & \cellcolor{BestCell}100.0 & \cellcolor{BestCell}100.0 & \cellcolor{BestCell}100.0 & \cellcolor{SecondCell}97.0 & \cellcolor{BestCell}97.0 \\
\midrule
\rowcolor{ModelHeader}\multicolumn{7}{l}{\textcolor{white}{\textbf{gpt-oss-20b (MXFP4)}}} \\
KEEP & \cellcolor{SecondCell}77.5 & \cellcolor{BestCell}59.3 & \cellcolor{ThirdCell}72.2 & \cellcolor{BestCell}66.7 & \cellcolor{SecondCell}82.1 & \cellcolor{SecondCell}63.9 \\
Always-COMPRESS & \cellcolor{ThirdCell}74.5 & \cellcolor{ThirdCell}48.1 & \cellcolor{ThirdCell}72.2 & \cellcolor{SecondCell}50.0 & 76.4 & 50.0 \\
Always-DROP & \cellcolor{BestCell}82.4 & 42.6 & \cellcolor{SecondCell}79.6 & \cellcolor{SecondCell}50.0 & \cellcolor{ThirdCell}79.7 & \cellcolor{BestCell}66.7 \\
RECENCY & \cellcolor{SecondCell}77.5 & \cellcolor{BestCell}59.3 & \cellcolor{ThirdCell}72.2 & \cellcolor{BestCell}66.7 & \cellcolor{SecondCell}82.1 & \cellcolor{SecondCell}63.9 \\
SEMANTIC & \cellcolor{SecondCell}77.5 & \cellcolor{SecondCell}55.6 & \cellcolor{BestCell}81.5 & \cellcolor{BestCell}66.7 & \cellcolor{BestCell}82.9 & \cellcolor{SecondCell}63.9 \\
LEARNED-NOCERT & \cellcolor{SecondCell}77.5 & \cellcolor{BestCell}59.3 & \cellcolor{ThirdCell}72.2 & \cellcolor{BestCell}66.7 & \cellcolor{SecondCell}82.1 & \cellcolor{ThirdCell}55.6 \\
\textbf{RECAP} & \cellcolor{SecondCell}77.5 & \cellcolor{BestCell}59.3 & \cellcolor{ThirdCell}72.2 & \cellcolor{BestCell}66.7 & \cellcolor{SecondCell}82.1 & \cellcolor{SecondCell}63.9 \\
\bottomrule
\end{tabular}
\end{table}

Overall continuation success need not equal an unweighted mean of the six
domain columns because checkpoint counts differ. Different backbones also
use different selected task pools, so these results are within-backbone
comparisons rather than a controlled scaling-law experiment.

\FloatBarrier
\section{Standalone Calibration and Additional Diagnostics}
\label{app:standalone-calibration}
\subsection{The separate 69-task Qwen3-8B calibration}
This calibration experiment contains 461 states and 4,149 continuations.
Its selected $(\lambda,\tau)=(0.36638,0)$ incurs five harmful-task and five
cost-failure outcomes among 69 tasks, giving $U_H=U_C=0.1611<0.20$.
The policy selects 109 edits (23.64\%; K/C/D $=352/19/90$), with seven
harmful states, and sums to 167,799 prompt tokens saved across the matched
continuations (363.99 per state). These are descriptions of the calibration
sample used for selection, not held-out performance or the saving of one
sequential episode. This certificate is separate from the backbone-specific
Qwen3-8B test row.

\begin{table}[htbp]
\centering
\caption{\textbf{Calibration outcomes for the stated task-bundle populations.}
One-sided tail probability is $\alpha/2=0.025$ per risk; budgets are both
$0.20$. The standalone Qwen3-8B study is not the 58-task calibration of the
five-backbone comparison. A dash denotes an unreported count.}
\label{tab:calibration}
\small
\setlength{\tabcolsep}{4.2pt}
\begin{tabular}{lrrrrrr}
\toprule
Calibration & Tasks & States & $k_H$ & $k_C$ & $U_H$ & $U_C$ \\
\midrule
Standalone Qwen3-8B & 69 & 461 & 5 & 5 & 0.1611 & 0.1611 \\
gpt-oss-20b & 30 & 87 & \textemdash & \textemdash & 0.1157 & 0.1722 \\
Hunyuan-A13B FP8 & 30 & 41 & 0 & 1 & 0.1157 & 0.1722 \\
\bottomrule
\end{tabular}
\end{table}

\subsection{Calibration-size sensitivity}
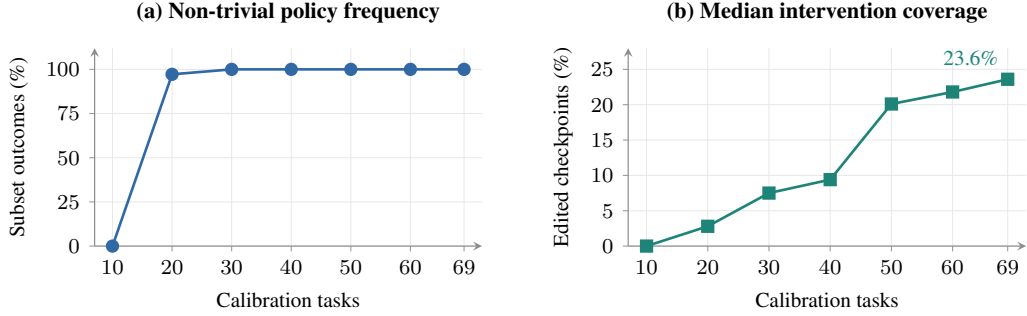
\begin{figure}[tbp]
\centering
\begin{minipage}[t]{0.48\linewidth}
\centering
\begin{tikzpicture}
\begin{axis}[
  recap/axis,height=4.2cm,title={(a) Non-trivial policy frequency},
  xmin=7,xmax=72,ymin=0,ymax=112,
  xtick={10,20,30,40,50,60,69},ytick={0,25,50,75,100},
  xlabel={Calibration tasks},ylabel={Subset outcomes (\%)}
]
\addplot[RECAPPlotBlue,mark=*]
 coordinates {(10,0) (20,97.2) (30,100) (40,100) (50,100) (60,100) (69,100)};
\end{axis}
\end{tikzpicture}
\end{minipage}\hfill
\begin{minipage}[t]{0.49\linewidth}
\centering
\begin{tikzpicture}
\begin{axis}[
  recap/axis,height=4.2cm,title={(b) Median intervention coverage},
  xmin=7,xmax=72,ymin=0,ymax=28,
  xtick={10,20,30,40,50,60,69},ytick={0,5,10,15,20,25},
  xlabel={Calibration tasks},ylabel={Edited checkpoints (\%)}
]
\addplot[RECAPPlotTeal,mark=square*]
 coordinates {(10,0) (20,2.8) (30,7.5) (40,9.4) (50,20.1) (60,21.8) (69,23.6)};
\node[anchor=south east,font=\fontsize{8}{9}\selectfont,text=RECAPPlotTeal]
 at (axis cs:69,24.2) {23.6\%};
\end{axis}
\end{tikzpicture}
\end{minipage}
\caption{\textbf{Calibration-size sensitivity in the standalone Qwen study.}
The existing analysis draws 500 task subsets at sizes below 69, with the
predictors and candidate order fixed; 69 uses the full pool.
Only the previously recorded frequencies and medians are plotted.
No uncertainty band is shown because repeat-level outputs are unavailable.
This is a diagnostic on an observed pool, not a new certificate or permission
to collect more tasks until a preferred policy passes.}
\label{fig:calibration-size}
\end{figure}
For calibration sizes $10,20,30,40,50,60,69$, the fractions of
non-trivial certified policies are $0,97.2,100,100,100,100,100\%$, and median
switch rates are $0,2.8,7.5,9.4,20.1,21.8,23.6\%$. This illustrates
finite-sample conservatism for the fixed family; it does not justify collecting
more calibration tasks until a preferred policy passes.

In the history-length diagnostic, Qwen3-14B has positive
paired-success savings for short/medium/long failure histories
(855/152/1,200 tokens), whereas Qwen3-8B has $-53/521/106$ tokens. These bins
use 2--3, 4--5, and at least 6 prior failures, respectively. The pattern is not
monotone. Additional history-length and statistical results appear in
Appendix~\ref{app:horizon}. Hunyuan has no recorded history-bin breakdown.

\subsection{Historical statistical and history-length reporting}
\label{app:horizon}

\SCIfFigureExists{figures/fig_horizon_stress}{%
\begin{figure}[ht]
\centering
\SCfigure[width=0.86\linewidth]{figures/fig_horizon_stress}
\caption{\textbf{Qwen history-length analysis.} Bins use the number of
failures already observed at the checkpoint. Results are paired-success token
savings; bins with fewer than ten eligible events are omitted. Hunyuan has no recorded history-bin breakdown.}
\label{fig:horizon-stress}
\end{figure}
}{%
\PackageInfo{RECAP}{Optional figure figures/fig_horizon_stress omitted: file not supplied}%
\paragraph{Qwen history-length analysis.}
Bins use the number of
failures already observed at the checkpoint. Results are paired-success token
savings; bins with fewer than ten eligible events are omitted. Hunyuan has no recorded history-bin breakdown.
}

\paragraph{Historical bootstrap summaries and present uncertainty.}
A preceding task-bootstrap analysis reports RECAP saving intervals of
$[-3{,}336,2{,}472]$ tokens for Qwen3-8B and $[280,1{,}479]$ for Qwen3-14B.
These historical summaries are kept separate: the current aggregate records
do not establish their exact correspondence to every policy, estimand, and
seed set in Tables~\ref{tab:main-multimodel} and~\ref{tab:main-crossfamily}. They therefore are not attached
to the current point estimates as verified confidence intervals. Hunyuan has
no recorded bootstrap interval. Neither a positive point estimate nor a
zero-event test sample establishes positive population benefit or zero harm.

For current uncertainty estimates, the resampling unit must preserve each
task's checkpoint--action--seed bundle and use the same task resample for all
compared selectors. Resampling individual checkpoints as independent data
would ignore their shared trajectory. A bootstrap of a zero-event sample may
degenerate to $[0,0]$, which must not be interpreted as zero population risk.
Task-level calibration bounds and held-out descriptive intervals answer
different questions.

\FloatBarrier
\section{Reproducibility, Splits, and Frozen Artifacts}
\label{app:reproducibility}

\subsection{Evidence map and version boundaries}
The main comparison follows one per-backbone chain: screened task split,
frozen matched K/C/D outcomes, development-fitted scores and ordered policies,
calibration task counts and bounds, then test selection from the held-out
matrix. The identifiers below distinguish this chain from the earlier
standalone and diagnostic studies.

\begin{table}[htbp]
\centering
\caption{\textbf{Which evidence supports which claim.} Studies are not pooled.
All quantities are retained from the reported records.}
\label{tab:evidence-map}
\small
\begin{tabularx}{\linewidth}{@{}p{0.24\linewidth}XX@{}}
\toprule
Study & Recorded unit & Supported interpretation \\
\midrule
Five-backbone comparison & Backbone-specific dev/cal/test tasks;
three seeds per checkpoint/action & Held-out single-edit operating points;
policy hashes in this appendix. \\
Fixed-block Qwen3-8B case & One task; seven checkpoints; five seeds;
105 continuations & Same $B$ and $B'$ across states;
not a prevalence estimate or learned-policy trace. \\
Development diagnostic & 39 tasks; 286 states; 2,574 rollouts &
Recovery-first preferences and predictor diagnostics;
not the formal token oracle. \\
Standalone Qwen calibration & 69 tasks; 461 states; 4,149 continuations &
A separate selected policy and calibration-size analysis;
not the main 58-task-calibrated Qwen test policy. \\
\bottomrule
\end{tabularx}
\end{table}

\subsection{Main-comparison splits}

\begin{table}[ht]
\centering
\caption{\textbf{Registered task counts and backbone-specific operating points.}
Hunyuan development/calibration/test have 14/15/15 tasks with retained states. The separate
69-task Qwen3-8B calibration study is not included in this table.}
\label{tab:splits}
\small
\setlength{\tabcolsep}{5pt}
\begin{tabular}{lrrrrr}
\toprule
Backbone & Dev. & Cal. & Test & $\lambda$ & $\tau$ \\
\midrule
Qwen3-4B & 35 & 44 & 21 & 0.291414 & 0.05 \\
Qwen3-8B & 35 & 58 & 40 & 0.060607 & 0 \\
Qwen3-14B AWQ & 31 & 57 & 41 & 0.134412 & 0 \\
gpt-oss-20b & 36 & 30 & 45 & 0.0004117 & 0.05 \\
Hunyuan-A13B FP8 & 24 & 30 & 19 & 0.0076923 & 0 \\
\bottomrule
\end{tabular}
\end{table}

The standalone 69-task Qwen3-8B calibration is a separate frozen experiment,
not the 58-task backbone-adaptation calibration in Table~\ref{tab:splits}.
For gpt-oss, development/calibration/test contain 104/87/129 states and
936/783/1,161 counterfactual rollouts. For Hunyuan, the corresponding counts are
32/41/37 states and 288/369/333 rollouts. The seven method selectors reuse
these continuations and are not seven new collections of model outputs.

\paragraph{Frozen policy hashes.}
The following line breaks are typographic only.
\begin{description}[leftmargin=1.4em,labelindent=0pt,itemsep=1pt,style=nextline]
\item[Standalone Qwen3-8B (69 tasks)] \texttt{888fe0f1\allowbreak{}c8e6246f\allowbreak{}9d1955f8\allowbreak{}2d2d4967\allowbreak{}ab88af7b\allowbreak{}a54779e4\allowbreak{}8d5c2394\allowbreak{}88517207}
\item[Qwen3-4B] \texttt{bcef0fb4\allowbreak{}d8dd2404\allowbreak{}2f4f3603\allowbreak{}391defc1\allowbreak{}bfaa6a35\allowbreak{}49dc6ee9\allowbreak{}290daf1d\allowbreak{}1fe1f700}
\item[Qwen3-8B] \texttt{16bda5e9\allowbreak{}e0b256e0\allowbreak{}fa69d29a\allowbreak{}cc7afa60\allowbreak{}45e15c25\allowbreak{}02cc7476\allowbreak{}ef3ae057\allowbreak{}99f800cf}
\item[Qwen3-14B AWQ] \texttt{d7e03cf9\allowbreak{}73171f09\allowbreak{}6cb4a990\allowbreak{}620c374a\allowbreak{}03f7e325\allowbreak{}386ab80c\allowbreak{}f77a8486\allowbreak{}c81cc287}
\item[gpt-oss-20b] \texttt{dac3c350\allowbreak{}aa7f6cdf\allowbreak{}cf6bde4f\allowbreak{}eeb83341\allowbreak{}44faf5da\allowbreak{}3c2df587\allowbreak{}6c5c376d\allowbreak{}2b597c59}
\item[Hunyuan-A13B FP8] \texttt{881aaabb\allowbreak{}41217d00\allowbreak{}c41b6366\allowbreak{}00d63256\allowbreak{}d4c0efc7\allowbreak{}b0ec2cca\allowbreak{}4e2b5dd0\allowbreak{}fdd03172}
\end{description}

\paragraph{Artifact provenance.}
Frozen policy hashes identify the standalone calibration study and each
backbone-specific policy. The seven-method evaluation records unchanged
policy hashes and reuses existing continuations. The relevant Hunyuan records
are \texttt{metrics.json}, \texttt{calibration\_candidate\_bounds.csv}, and
\texttt{baseline\_summary.json}, at
\path{outputs/safe_cure/cross_family_supported/hunyuan/metrics.json},
\path{outputs/safe_cure/cross_family_supported/hunyuan/calibration_candidate_bounds.csv},
and
\path{outputs/safe_cure/cross_family_supported/strong_baselines/baseline_summary.json}.
The legacy directory names are retained for traceability. Exact replication
also requires task manifests, model revisions, serving configuration,
task-generation and rollout seeds, per-seed outcomes, and the implementation
specifications identified in Appendix~\ref{app:replay}.

\FloatBarrier
\section{Extended Related Work}
\label{app:related-work}
\paragraph{Prompt compression and long-context use.}
Longer input windows do not ensure effective use of retained information:
performance can depend strongly on where relevant evidence appears
\citep{liu2024lost}. Selective Context removes low-information content
\citep{li2023selectivecontext}; LLMLingua and LongLLMLingua compress prompts
using token importance and question-aware processing
\citep{jiang2023llmlingua,jiang2024longllmlingua}. LLMLingua-2 learns
extractive compression through data distillation \citep{pan2024llmlingua2}.
Gist tokens and the In-context Autoencoder instead learn compact latent
representations \citep{mu2023gist,ge2024icae}. These approaches concern how to
represent context economically. RECAP studies a complementary decision:
when to retain, summarize, or remove a selected failure block, evaluated by
matched continuation outcomes rather than compression ratio alone.

\paragraph{Persistent memory and experience reuse.}
Generative Agents combines experience storage, reflection, and retrieval
\citep{park2023generative}; MemoryBank updates memories with time- and
significance-dependent retention \citep{zhong2024memorybank}; MemGPT manages
information across memory tiers \citep{packer2023memgpt}. A-MEM dynamically
links and revises structured notes \citep{xu2025amem}, while Mem0 extracts,
consolidates, and retrieves salient conversational information
\citep{chhikara2025mem0}. Sentence embeddings support similarity-based
retrieval \citep{reimers2019sentencebert}. Feedback also guides future
behavior: ReAct interleaves reasoning and action, Reflexion retains verbal
feedback, Self-Refine iteratively revises outputs, and Voyager accumulates
executable skills \citep{yao2022react,NEURIPS2023_1b44b878,madaan2023selfrefine,wang2023voyager}.
These works organize or reuse experience. RECAP complements them by measuring
the continuation-level effect of changing one identified failure block at a
fixed execution checkpoint.

\paragraph{Learned context operations and counterfactual selection.}
Context-Folding learns to branch into sub-trajectories and summarize completed
subtasks \citep{sun2025contextfolding}; AgentFold learns multi-scale trajectory
condensation \citep{ye2025agentfold}; AgeMem learns memory operations including
storage, retrieval, updating, summarization, and discarding
\citep{yu2026agemem}. Closely related, ASSAY estimates per-skill causal
contributions through randomized masking and suppresses skills with negative
predicted effects for a test task \citep{wang2026assay}. Thus, both learned
forgetting and counterfactual knowledge selection have precedents. RECAP
focuses on failure-block edits within an ongoing execution, compares KEEP,
COMPRESS, and DROP from matched checkpoints, and separately certifies two
policy-level task-bundle risks. End-to-end memory and skill-management policies
require separate rollouts and cannot be compared through our frozen
single-checkpoint action matrix.

\paragraph{Selective prediction and statistical risk control.}
Selective classification trades coverage for prediction risk through
abstention \citep{geifman2017selective}. Risk-controlling prediction sets use
held-out calibration to obtain risk guarantees \citep{bates2021rcps}, while
Conformal Risk Control controls the expected value of a monotone loss
\citep{angelopoulos2024crc}. Prompt Risk Control selects LLM prompts using
upper bounds on specified risk measures \citep{zollo2024promptrisk}.
RECAP uses Learn-Then-Test \citep{angelopoulos2021learntest} with one-sided
Clopper--Pearson bounds \citep{clopper1934binomial} for harmful-intervention
and non-positive-benefit losses. KEEP is abstention relative to the editing
decision, not a claim that the underlying agent cannot fail. Certification
selects from a development-fixed policy sequence; it neither assumes that
the two-dimensional threshold grid is nested nor certifies every edit.

\paragraph{Baseline-relative policy improvement.}
High-confidence off-policy evaluation bounds the value of a policy using
data collected by other policies \citep{thomas2015hcope}. Safe Policy
Improvement with Baseline Bootstrapping retains baseline behavior where
offline evidence is insufficient \citep{laroche2019spibb}. RECAP shares the
motivation of controlling deviations from a reference, but directly evaluates
all three context treatments at retained checkpoints. It does not estimate
full-policy returns from an observational action log, and its task-bundle
certificate is not a guarantee for repeated edits in an evolving episode.

\paragraph{Evaluation of context, memory, and interactive agents.}
LongBench and RULER evaluate long-context understanding and effective context
use \citep{bai2024longbench,hsieh2024ruler}; LoCoMo evaluates very long-term
conversational memory \citep{maharana2024locomo}. AgentBench, WebArena, and
$\tau$-bench evaluate multi-step interaction, web tasks, and tool--agent--user
interaction, respectively \citep{liu2024agentbench,zhou2024webarena,yao2024taubench}.
Our experiments use the six domains of AgentCE-Bench, whose lightweight
planning environments provide configurable horizons and difficulty
\citep{yang2026agentce}. We evaluate RECAP only on the screened, retained
AgentCE-Bench checkpoints specified in Section~\ref{sec:problem}; the other
benchmarks motivate complementary evaluation settings.

\end{document}